\documentclass[11pt,letterpaper]{article}
\usepackage{hunyuan-report}
\usepackage{amsmath,amsfonts,bm}

\def\eqref#1{equation~\ref{#1}}

\def\1{\bm{1}}

\DeclareMathAlphabet{\mathsfit}{\encodingdefault}{\sfdefault}{m}{sl}
\SetMathAlphabet{\mathsfit}{bold}{\encodingdefault}{\sfdefault}{bx}{n}

\usepackage{hyperref}
\usepackage{url}
\usepackage{booktabs}
\usepackage{graphicx}
\usepackage{multirow}
\usepackage{makecell}
\usepackage{listings}
\usepackage[most]{tcolorbox}

\lstdefinestyle{promptstyle}{
basicstyle=\ttfamily\scriptsize,
breaklines=true,
breakatwhitespace=false,
columns=fullflexible,
keepspaces=true,
showstringspaces=false,
frame=none,
aboveskip=0pt,
belowskip=0pt
}

\newtcblisting{promptbox}[1]{
breakable,
listing only,
listing options={style=promptstyle},
title=\textbf{#1},
colback=gray!3,
colframe=gray!45,
colbacktitle=gray!10,
coltitle=black,
boxrule=0.5pt,
arc=1pt,
left=5pt,
right=5pt,
top=4pt,
bottom=4pt,
before skip=6pt,
after skip=8pt
}

\hypersetup{
  pdftitle={Scaling Unified Multimodal Models to Long-Horizon Interleaved Generation},
  pdfauthor={Ziqiao Peng; Yutao Cui; Wang Lin; Zhong-Yu Li; Qingwen Gu; Tianhui Song; Chengyou Jia; Guozhen Zhang; Xuerui Qiu; Qixun Wang; Miles Yang; Zhao Zhong; Yonglong Tian; Hongyan Liu; Jun He},
  colorlinks=true,linkcolor=HYBlueDark,citecolor=HYBlueDark,urlcolor=HYBlue
}
\ifPDFTeX\pdfminorversion=7\else\fi
\begin{document}
\thispagestyle{firstpage}
\begingroup
\setlength{\parskip}{0pt}
\noindent\includegraphics[width=125pt]{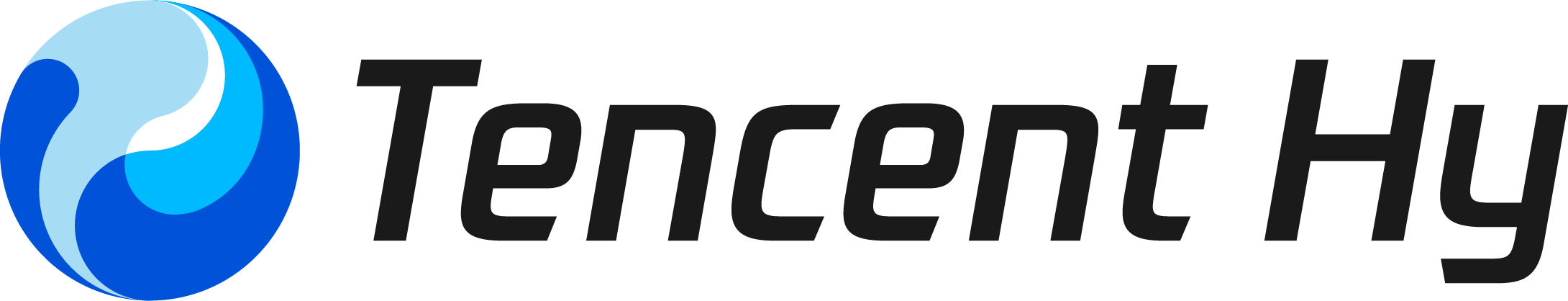}
\par\vspace{3pt}
{\color{HYBlue}\hrule height 1.2pt}
\vspace{5pt}
{\centering
{\sffamily\bfseries\fontsize{20}{23}\selectfont
Scaling Unified Multimodal Models to\\[2pt]
Long-Horizon Interleaved Generation\par}
\vspace{4pt}
{\color{HYBlue!40}\hrule height 0.4pt}
\vspace{6pt}
{\authorface\fontsize{10}{13}\selectfont
Ziqiao Peng\textsuperscript{1,*}\quad
Yutao Cui$^{2,\dagger}$\quad
Wang Lin\textsuperscript{2}\quad
Zhong-Yu Li\textsuperscript{2}\quad
Qingwen Gu\textsuperscript{2}\\
Tianhui Song\textsuperscript{2}\quad
Chengyou Jia\textsuperscript{2}\quad
Guozhen Zhang\textsuperscript{2}\quad
Xuerui Qiu\textsuperscript{2}\quad
Qixun Wang\textsuperscript{2}\\
Miles Yang\textsuperscript{2}\quad
Zhao Zhong\textsuperscript{2}\quad
Yonglong Tian\textsuperscript{2}\quad
Hongyan Liu$^{3,\ddagger}$\quad
Jun He$^{1,\ddagger}$\par}
\vspace{4pt}
{\fontsize{9}{11}\selectfont
\textsuperscript{1}Renmin University of China\quad
\textsuperscript{2}Tencent Hy\quad
\textsuperscript{3}Tsinghua University\par}
\vspace{4pt}
{\fontsize{8}{10}\selectfont\color{HYGray}
\textsuperscript{*}Work done during an internship at Tencent Hy.\\[2pt]
\textsuperscript{\textdagger}Project Lead.\qquad
\textsuperscript{\textdaggerdbl}Corresponding Authors.\par}
}
\endgroup
\vspace{5pt}
\begin{tcolorbox}[colback=white,colframe=HYBlue!55,boxrule=0.45pt,
arc=3pt,left=11pt,right=11pt,top=6pt,bottom=8pt]
{\centering\sffamily\bfseries\fontsize{12}{14}\selectfont\color{HYBlueDark}Abstract\par}\vspace{4pt}
{\fontsize{10}{12.2}\selectfont Unified multimodal models (UMMs) provide an interface for interleaved text-image generation, where generated visual states can directly inform subsequent reasoning and synthesis. However, existing systems are still primarily demonstrated on short trajectories or simplified visual content. We develop a large UMM natively trained on interleaved multimodal data, with strong capabilities in image synthesis, text rendering, complex composition, and multimodal planning, and study how to extend these native capabilities to substantially longer autoregressive trajectories. The main challenge is that generated images continuously accumulate in context: irrelevant or imperfect visual states interfere with future synthesis, while previously established characters and scenes must still be recovered after long intervals. We introduce a visual-memory framework that maintains persistent visual references, selectively retrieves scene-relevant historical states while preserving the complete semantic history, and uses few-step self-rollout to adapt the model to imperfect generated histories. We further introduce LongMIG-Bench for evaluating long-horizon text-to-multi-image and image-to-multi-image generation. Our model achieves high-fidelity long-horizon interleaved generation with strong character and scene consistency, demonstrating how visual memory can scale native interleaved generation to substantially longer trajectories.
\par}
\end{tcolorbox}

\begin{figure}[p]
\begin{center}
\includegraphics[width=\linewidth]{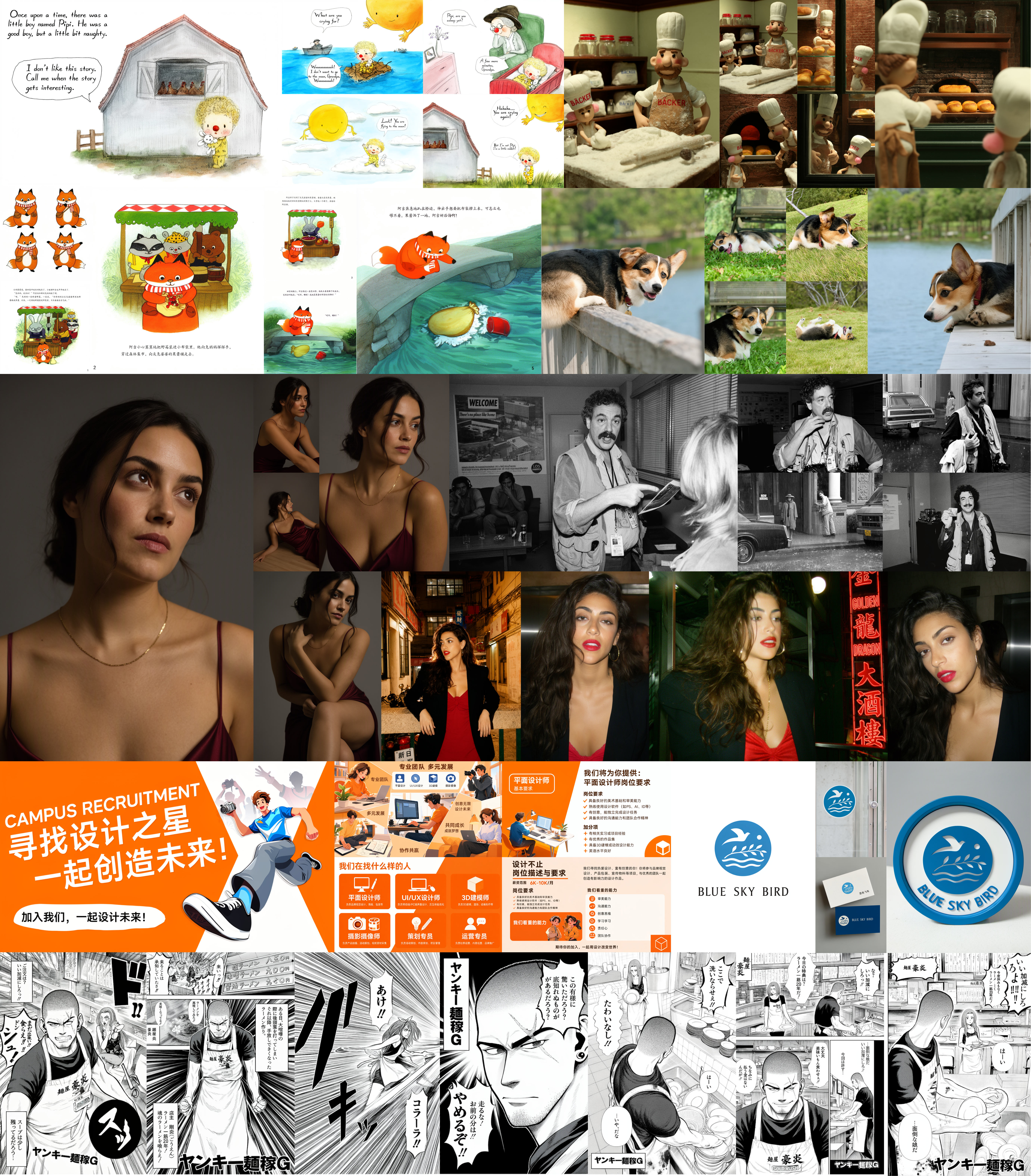}
\end{center}

\caption{\textbf{High-fidelity long-horizon interleaved generation across diverse visual domains.} Our model supports photorealistic, stylized, text-rich, and design-oriented generation while maintaining visual consistency over extended trajectories.}

\label{fig:release-teaser}
\end{figure}

\section{Introduction}

Unified multimodal models (UMMs) are evolving from systems that separately understand or generate visual content into general-purpose models that can reason and generate across text and images within a single autoregressive trajectory~\citep{team2024chameleon,deng2025bagel,wu2025janus,wang2024emu3,wang2026illuminating,xie2026show}. Native interleaving provides a natural formulation for long-form multimodal generation: generated visual states can directly inform subsequent reasoning and synthesis, allowing the model to continuously update its plan based on previously generated text and images rather than reducing each step to an independent prompt. It also provides a natural interface for learning from sequential visual data such as videos, comics, storyboards, tutorials, and other multimodal sequences. These properties make native interleaved generation a promising paradigm for applications including illustrated stories, visual design, tutorials, and multi-shot content creation.

Despite this potential, existing interleaved systems are still primarily demonstrated on relatively short trajectories or simplified visual content~\citep{yang2025seed,chen2026long}. In this work, we train a large UMM natively on mixed text-image trajectories, where visual outputs remain within the shared multimodal sequence and can directly condition subsequent reasoning and synthesis. The resulting model supports high-fidelity image synthesis, text rendering, complex composition, and multimodal planning within a single autoregressive trajectory. Our goal is not merely to generate more images in sequence, but to preserve these native capabilities over substantially longer horizons.

However, native interleaved training alone does not resolve the challenges that emerge as the trajectory becomes long. Generated images themselves become part of the future context, and every image introduces a dense visual state that may influence later synthesis. Some states remain essential, such as the appearance of a recurring character or an earlier view of a scene, while others become redundant, irrelevant, or imperfect. Conditioning every image on the complete visual history therefore leads to continuously growing visual context and increasing interference, while relying only on recent images may discard information needed for long-range revisits~\citep{ye2026loom}.

Recent work has similarly shown that retaining more visual context does not necessarily improve long-horizon generation~\citep{chen2026long}. Meanwhile, teacher-forced training conditions on clean ground-truth histories, whereas autoregressive inference must continue from the model's own outputs, allowing small visual errors to accumulate over time~\citep{huang2026self,chen2024diffusion}.

We address these challenges through an explicit visual-memory framework. The model first forms persistent visual references for information that should remain stable throughout the trajectory. Before each image generation, its understanding pathway predicts the current scene state and selectively retrieves relevant historical visual states. Importantly, this selection applies only to dense generative visual representations, while the complete semantic history remains available for global reasoning. We further introduce few-step self-rollout adaptation, which exposes the model during post-training to its own generated histories while retaining ground-truth targets for supervision. These mechanisms reduce unnecessary historical visual conditioning while preserving information required for long-range consistency.

To evaluate this setting, we introduce LongMIG-Bench for long-horizon text-to-multi-image and image-to-multi-image generation. The benchmark contains recurring characters, scene transitions, and long-range revisits, requiring models to preserve and recover visual information across multiple intervening generations. Experiments show that our model achieves high-fidelity long-horizon interleaved generation with strong character and scene consistency, while controlled analyses demonstrate the complementary roles of persistent visual references, scene-aware retrieval, and self-rollout adaptation. Our memory design also keeps the dense historical visual context used for image synthesis compact, enabling longer and more efficient autoregressive trajectories.

\Needspace{8\baselineskip}
Our contributions are threefold:
\begin{itemize}
    \item We develop a large UMM natively trained for interleaved text-image generation and study how to scale its image synthesis, text rendering, complex composition, and multimodal planning capabilities to long autoregressive trajectories.
    \item We develop a visual-memory framework that combines persistent visual references, scene-aware historical retrieval, and few-step self-rollout adaptation, while separating complete semantic history from the dense visual states directly used for image synthesis.
    \item We introduce LongMIG-Bench to evaluate long-horizon interleaved generation and demonstrate high-fidelity generation with strong long-range character and scene consistency.
\end{itemize}

\section{Related Work}

\subsection{Unified Multimodal and Long-Horizon Generation}

Recent unified multimodal models~\citep{diao2026sensenova,xie2026show} integrate visual understanding and generation within a single architecture, enabling text and images to be processed and generated in flexible interleaved sequences. 
Unlike agent-based pipelines~\citep{wang2024genartist, zhang2026qwen} that reduce each generation step to an independent image prompt, native interleaved models retain generated visual states within the same autoregressive trajectory, enabling generation and planning to continuously condition on previous multimodal outputs.
Representative models such as Janus~\citep{wu2025janus}, BAGEL~\citep{deng2025bagel}, Show-o~\citep{xie2025show}, SEED-Story~\citep{yang2025seed}, Wan-Weaver~\citep{xing2026wan}, and ChatUMM~\citep{dai2026chatumm} demonstrate increasingly strong capabilities in multimodal understanding, image generation, and interleaved text-image generation. These advances establish that unified models can produce multimodal outputs within a single generative trajectory.

Our focus is on how this capability scales to substantially longer horizons. As generated images accumulate in the context, models must preserve visual information across distant steps while controlling an increasingly large history. Related problems have been studied in long-context modeling through selective state retention and KV-cache compression~\citep{xiao2024efficient,li2024snapkv}. More closely, UniLongGen~\citep{chen2026long} shows that retaining the complete visual history can degrade long-form generation and introduces training-free context curation to remove less relevant historical states. Our work shares the motivation of avoiding indiscriminate full-history conditioning, but instead organizes visual history as explicit memory: the model preserves the full semantic trajectory for global reasoning while selectively exposing task-relevant visual states to image generation. We therefore study visual-memory management as a mechanism for scaling native interleaved generation to long multimodal trajectories.

\subsection{Visual Memory and Self-Conditioned Generation}

Recent multimodal models have begun to treat generated images as intermediate reasoning states. Works on generated visual thoughts and multimodal chain-of-thought~\citep{chern2025thinking,qin2026uni,chen2026unit,yin2026generative,hu2026tvi} show that visual outputs can support reasoning, refinement, and test-time computation. Our setting differs in that generated visual states are used as \emph{persistent memory} rather than transient intermediate reasoning. The model autonomously creates reusable character references because their information will be needed repeatedly later in the trajectory. This is also related to subject-driven generation~\citep{ruiz2023dreambooth,chen2023subject,ye2023ip}, but our references are generated by the model itself rather than supplied externally.

Long-horizon generation amplifies the train--test mismatch of autoregressive generation. Teacher-forced training conditions on ground-truth histories, whereas inference must continue from the model's own imperfect outputs. Rollout-based approaches such as Self-Forcing~\citep{huang2026self}, Self-Forcing++~\citep{cui2026self}, and Rolling Forcing~\citep{liu2026rolling} address exposure gaps in autoregressive video generation. We study the analogous problem for interleaved multimodal sequences, using few-step self-rollout to expose the model to self-generated visual histories while retaining ground-truth targets for supervision. These directions motivate our view of long-horizon interleaved generation as a visual-memory problem: models must determine what visual information should persist, what should be retrieved at each step, and how to continue from imperfect history.

\begin{figure}[t]
\begin{center}
\includegraphics[width=\linewidth]{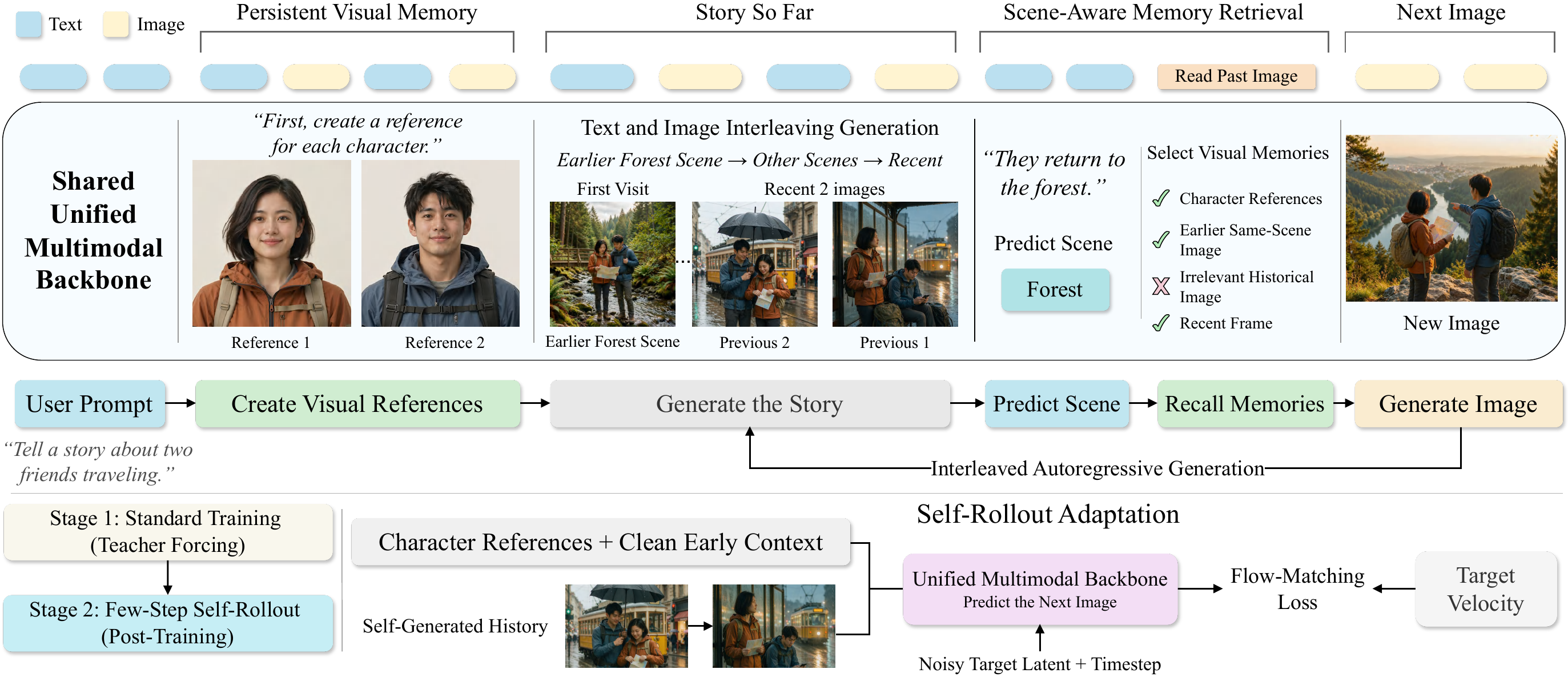}
\end{center}
\vspace{-12pt}
\caption{\textbf{Overview of our long-horizon interleaved generation framework.} Built on our unified multimodal backbone, the model constructs persistent visual references, retrieves scene-relevant history during autoregressive generation, and uses few-step self-rollout for memory adaptation.}
\vspace{-10pt}
\end{figure}

\section{Method}
\label{sec:method}

\subsection{Overview}
\label{sec:overview}

We consider long-horizon interleaved generation with our unified multimodal model, natively trained on mixed text-image trajectories. During training, text and image outputs are represented within the same evolving multimodal sequence, allowing subsequent reasoning and synthesis to condition directly on preceding multimodal states. Given an instruction $x$, the model produces
\begin{equation}
Y = (y_1,y_2,\ldots,y_T),
\end{equation}
where each $y_t$ is either text or an image. Let $\tau_k$ denote the position of the $k$-th generated image $I_k$ in the multimodal sequence. Native interleaved generation then conditions on the preceding multimodal history,
\begin{equation}
p_\theta(I_k \mid x,Y_{<\tau_k}).
\end{equation}

Native interleaved training provides the model with the ability to reason and generate within a shared evolving multimodal trajectory, but does not by itself resolve the accumulation of dense visual states at long horizons. As generation proceeds, each previous image contributes a dense visual state, although only a subset may remain useful for the current synthesis.

We formulate long-horizon interleaved generation as a visual-memory problem. For each historical image $I_j$, the model maintains two representations: a semantic representation $h_j^{\mathrm{sem}}$ for multimodal understanding and a generative visual representation $m_j$ for image synthesis. We preserve the semantic trajectory for global reasoning, while controlling which historical generative representations condition each new image. This allows the model to reason over the full multimodal history without requiring every previous image to participate in visual generation. Our framework has three components: persistent visual memory for stable information, scene-aware retrieval for selecting relevant historical states, and self-rollout adaptation for training on imperfect autoregressive histories.

\subsection{Persistent Visual Memory}
\label{sec:persistent_memory}

Long-form generation repeatedly revisits information that should remain stable despite changes in pose, viewpoint, lighting, or composition. Character identity is a representative example. Relying only on arbitrary previously generated story images makes such information increasingly difficult to recover as the trajectory grows, since those images may themselves contain pose-specific details or generation errors.

We therefore allow the model to explicitly construct reusable visual references before long-form generation. Given instruction $x$, the model first generates a multimodal reasoning trajectory containing a set of visual anchors
\begin{equation}
\mathcal{A}=\{A_1,\ldots,A_M\},
\end{equation}
followed by the interleaved output trajectory,
\begin{equation}
\underbrace{\langle\mathrm{think}\rangle\, r^{\mathrm{text}}, A_1,\ldots,A_M\,\langle/\mathrm{think}\rangle}_{\text{visual memory formation}}
\quad
\underbrace{\langle\mathrm{answer}\rangle\, t_1,I_1,\ldots,t_N,I_N\,\langle/\mathrm{answer}\rangle}_{\text{interleaved generation}}.
\end{equation}
The anchors are generated by the model itself rather than provided externally and are trained using the model's native image-generation objective. Once generated, their clean generative representations are retained throughout the trajectory. They therefore provide stable visual references that can be reused at distant generation steps instead of repeatedly reconstructing identity from imperfect intermediate outputs.

\subsection{Scene-Aware Visual Memory Retrieval}
\label{sec:memory_retrieval}

Persistent anchors preserve globally important information, but ordinary historical images are not equally useful for every generation step. A fixed recent window preserves local continuity but fails when the model revisits an earlier scene, whereas full-history conditioning introduces an increasing number of irrelevant visual states. We therefore use the model's understanding pathway to determine which historical images should directly participate in image synthesis.

Before generating $I_k$, the model predicts a discrete scene state
\begin{equation}
s_k \in \{0,1,\ldots,S-1\},
\end{equation}
from the instruction and preceding multimodal trajectory. During training, scene prediction is supervised using scene annotations; during inference, $s_k$ is produced autoregressively by the model itself. The predicted state acts as an interface between multimodal understanding and visual generation: the model first determines the semantic state of the upcoming image and then retrieves visual memories associated with that state.

For generation step $k$, we construct the accessible generative visual memory as
\begin{equation}
\mathcal{M}_k=\mathcal{A}\cup\mathcal{R}_k\cup\mathcal{F}_k\cup\mathcal{L}_k,
\end{equation}
where $\mathcal{A}$ contains persistent visual anchors, $\mathcal{R}_k$ contains the most recent $K_r$ visual states regardless of scene, $\mathcal{F}_k$ contains early memories from the current scene, and $\mathcal{L}_k$ contains its most recent memories. For an ordinary historical image $I_j$, visibility to the generation of $I_k$ can be written as
\begin{equation}
g(k,j)=
\mathbb{I}[0<k-j\leq K_r]
\;\vee\;
\Big(
\mathbb{I}[s_j=s_k]
\wedge
\big[
\mathbb{I}[\rho(j)<N]
\vee
\mathbb{I}[\rho(k)-M\leq\rho(j)<\rho(k)]
\big]
\Big),
\end{equation}
where $\rho(j)$ denotes the occurrence index of $I_j$ within its scene. The recent window maintains short-range continuity, while scene-aware retrieval recovers both establishing and recent views when a scene is revisited after a long interval.

Crucially, this routing is applied only to historical generative visual representations. Images excluded from $\mathcal{M}_k$ remain accessible through their semantic representations $h_j^{\mathrm{sem}}$, allowing the understanding pathway to reason over the complete trajectory. We implement the retrieval rule by controlling attention from current image-generation queries to historical visual-memory keys, so excluded visual blocks need not participate in dense generative attention. The resulting generative visual memory therefore remains small even as the semantic trajectory continues to grow.

\subsection{Self-Rollout Adaptation}
\label{sec:self_rollout}

Memory formation and retrieval determine what information is available to the generator, but long autoregressive trajectories introduce an additional train--test mismatch. Standard teacher-forced training conditions each image on clean ground-truth history,
\begin{equation}
I_1^*,I_2^*,\ldots,I_{k-1}^* \rightarrow I_k^*,
\end{equation}
whereas inference conditions on the model's own previous generations,
\begin{equation}
\hat I_1,\hat I_2,\ldots,\hat I_{k-1}\rightarrow\hat I_k.
\end{equation}
Small deviations in identity, color, structure, or composition therefore become part of the subsequent visual context and may accumulate over many generation steps.

We reduce this exposure gap with a lightweight self-rollout stage. Starting from the teacher-forced model, we generate short autoregressive trajectories and replace a subset of historical ground-truth images with model-generated outputs. For selected rollout positions $\mathcal{S}_{\mathrm{roll}}$, we define
\begin{equation}
\widetilde I_j=
\begin{cases}
\hat I_j, & j\in\mathcal{S}_{\mathrm{roll}},\\
I_j^*, & \text{otherwise},
\end{cases}
\end{equation}
and train the subsequent image using
\begin{equation}
p_\theta(I_k^* \mid x,\widetilde I_{<k}).
\end{equation}
The generated images are used only as historical conditions, while the supervision target remains the original ground-truth image. This exposes the model to realistic inference-time histories without directly treating its previous errors as targets. In practice, we retain persistent visual anchors and a small amount of clean early context while rolling out several later visual states, keeping the adaptation stage lightweight.

\section{Experiments}
\label{sec:experiments}

\subsection{Experimental Setup}
\label{sec:exp_setup}

\paragraph{Model and training.}
Our backbone is a large-scale Mixture-of-Transformers (MoT) model, comparable in scale to recent large Hunyuan image-generation models, with separate understanding and generation branches using sparse expert activation. We natively train the model on diverse single-image and interleaved multimodal data, jointly learning multimodal understanding, autoregressive text generation, and high-fidelity image synthesis within a shared multimodal trajectory. The model supports multimodal contexts of up to 256K tokens.

\paragraph{Benchmark.}
We evaluate on \textbf{LongMIG-Bench}, which contains 100 stories evenly split between text-to-multi-image (\textsc{T2MI}) and image-to-multi-image (\textsc{I2MI}) generation, comprising 462 and 460 specified shots, respectively. Each story includes structured character, scene, and shot specifications, with recurring scenes designed to test visual consistency beyond a short recency window. 

\paragraph{Baselines.}
We compare two groups of methods according to their evaluation protocol.
For \textit{Agent-Planned} generation, an external planner decomposes each story into per-image prompts, which are used to evaluate StoryDiffusion~\citep{zhou2024storydiffusion}, Qwen-Image~\citep{zhao2026qwen}, UniSpace~\citep{yan2026unispace}, HiDream~\citep{cai2026hidream}, Lance-3B~\citep{fu2026lance}, and Show-o2~\citep{xie2026show}.
For native \textit{Interleaved} generation, we evaluate BAGEL~\citep{deng2025bagel}, SenseNova-U1.5~\citep{diao2026sensenova}, ILLUME-X~\citep{wang2026illuminating}, Emu3.5~\citep{cui2025emu3}, and SEED-Story~\citep{yang2025seed}.
We additionally reimplement UniLongGen~\citep{chen2026long} on our backbone, denoted as UniLongGen$^{*}$, for a controlled same-backbone comparison.

\paragraph{Evaluation.}
We use GPT-6 to evaluate \textbf{Character Identity}, \textbf{Scene Memory}, \textbf{Visual Fidelity}, \textbf{Visual Memory}, \textbf{Aesthetic Quality}, and \textbf{Text Rendering} on a 1--5 scale. Visual Memory specifically measures whether visual decisions introduced by earlier generated images persist later in the trajectory. We additionally report SceneGap using SigLIP2~\citep{tschannen2025siglip} (\textbf{Gap-S}) and DINOv3~\citep{simeoni2025dinov3} (\textbf{Gap-D}), defined as the difference between average same-scene and cross-scene similarity.

\subsection{Main Results}
\label{sec:main_results}

\begin{table*}[t]
\centering
\caption{\textbf{Long-horizon generation on LongMIG-Bench.} Best and second-best results are shown in \textbf{bold} and \underline{underline}, respectively.}
\label{tab:main_results}
\small
\setlength{\tabcolsep}{3.0pt}
\resizebox{\textwidth}{!}{
\begin{tabular}{lcccccccc}
\toprule
\multirow{2}{*}{\textbf{Method}}
& \multirow{2}{*}{\begin{tabular}[c]{@{}c@{}}\textbf{Character}\\\textbf{Identity} $\uparrow$\end{tabular}}
& \multicolumn{3}{c}{\textbf{Scene Memory}}
& \multirow{2}{*}{\begin{tabular}[c]{@{}c@{}}\textbf{Visual}\\\textbf{Fidelity} $\uparrow$\end{tabular}}
& \multirow{2}{*}{\begin{tabular}[c]{@{}c@{}}\textbf{Visual}\\\textbf{Memory} $\uparrow$\end{tabular}}
& \multirow{2}{*}{\begin{tabular}[c]{@{}c@{}}\textbf{Aesthetic}\\\textbf{Quality} $\uparrow$\end{tabular}}
& \multirow{2}{*}{\begin{tabular}[c]{@{}c@{}}\textbf{Text}\\\textbf{Rendering} $\uparrow$\end{tabular}} \\
\cmidrule(lr){3-5}
& & \textbf{Scene} $\uparrow$ & \textbf{Gap-S} $\uparrow$ & \textbf{Gap-D} $\uparrow$ & & & & \\
\midrule
\multicolumn{9}{l}{\textit{Agent-Planned}} \\
StoryDiffusion & 1.61 & 1.61 & 0.0615 & 0.1196 & 2.92 & 1.55 & 2.31 & 1.17 \\
Qwen-Image & 3.16 & 3.05 & 0.0694 & \underline{0.2127} & \textbf{3.83} & 3.14 & \underline{3.73} & 2.22 \\
UniSpace & 2.41 & 2.33 & 0.0346 & 0.1321 & 2.48 & 2.03 & 2.57 & 1.53 \\
HiDream & 3.35 & 2.58 & 0.0576 & 0.1620 & 3.19 & 2.76 & 2.71 & 2.78 \\
Lance-3B & 2.72 & 2.13 & 0.0480 & 0.1141 & 2.98 & 1.94 & 2.85 & 2.17 \\
Show-o2-7B & 2.72 & 2.90 & \underline{0.0783} & 0.1966 & 2.89 & 2.37 & 2.83 & 1.00 \\
\midrule
\multicolumn{9}{l}{\textit{Interleaved}} \\
BAGEL-7B & 3.38 & 1.44 & -0.0099 & -0.0191 & 2.84 & 1.70 & 2.52 & 1.35 \\
SenseNova-U1.5 & \textbf{3.72} & \underline{3.10} & 0.0062 & 0.0280 & 3.63 & \underline{3.49} & 3.30 & \underline{3.13} \\
ILLUME-X & 2.80 & 1.43 & -0.0098 & -0.0173 & 2.29 & 1.96 & 2.13 & 1.28 \\
Emu3.5 & 3.51 & 2.97 & 0.0350 & 0.1077 & 3.73 & 3.24 & 3.69 & 2.15 \\
SEED-Story & 1.37 & 1.14 & -0.0209 & -0.0421 & 2.91 & 2.93 & 1.70 & 2.43 \\
UniLongGen$^{*}$ & 3.46 & 2.61 & -0.0081 & -0.0135 & 3.05 & 2.60 & 2.97 & 2.28 \\
\textbf{Ours} & \underline{3.62} & \textbf{3.35} & \textbf{0.0930} & \textbf{0.2499} & \underline{3.75} & \textbf{3.77} & \textbf{3.75} & \textbf{3.50} \\
\bottomrule
\end{tabular}}
\end{table*}

Table~\ref{tab:main_results} shows that our model combines strong long-range visual memory with high per-image generation quality. Ours achieves the best Scene score, Gap-S, Gap-D, Visual Memory, Aesthetic Quality, and Text Rendering among all evaluated systems, while ranking second in Character Identity and Visual Fidelity. In particular, the strong Visual Memory score indicates that the model preserves visual decisions introduced by earlier generated frames, rather than reconstructing each image independently from the textual specification. The same-backbone comparison with UniLongGen$^{*}$ further isolates the effect of visual-memory management. Under the same underlying model, our method improves Scene from $2.61$ to $3.35$, Visual Memory from $2.60$ to $3.77$, and Visual Fidelity from $3.05$ to $3.75$, while changing SceneGap from negative values to strongly positive separation under both SigLIP2 and DINOv3. This controlled comparison suggests that explicit memory formation and scene-aware retrieval provide benefits beyond training-free context curation.

\begin{figure*}[t]
\centering
\makebox[\textwidth][c]{    \includegraphics[width=1\textwidth]{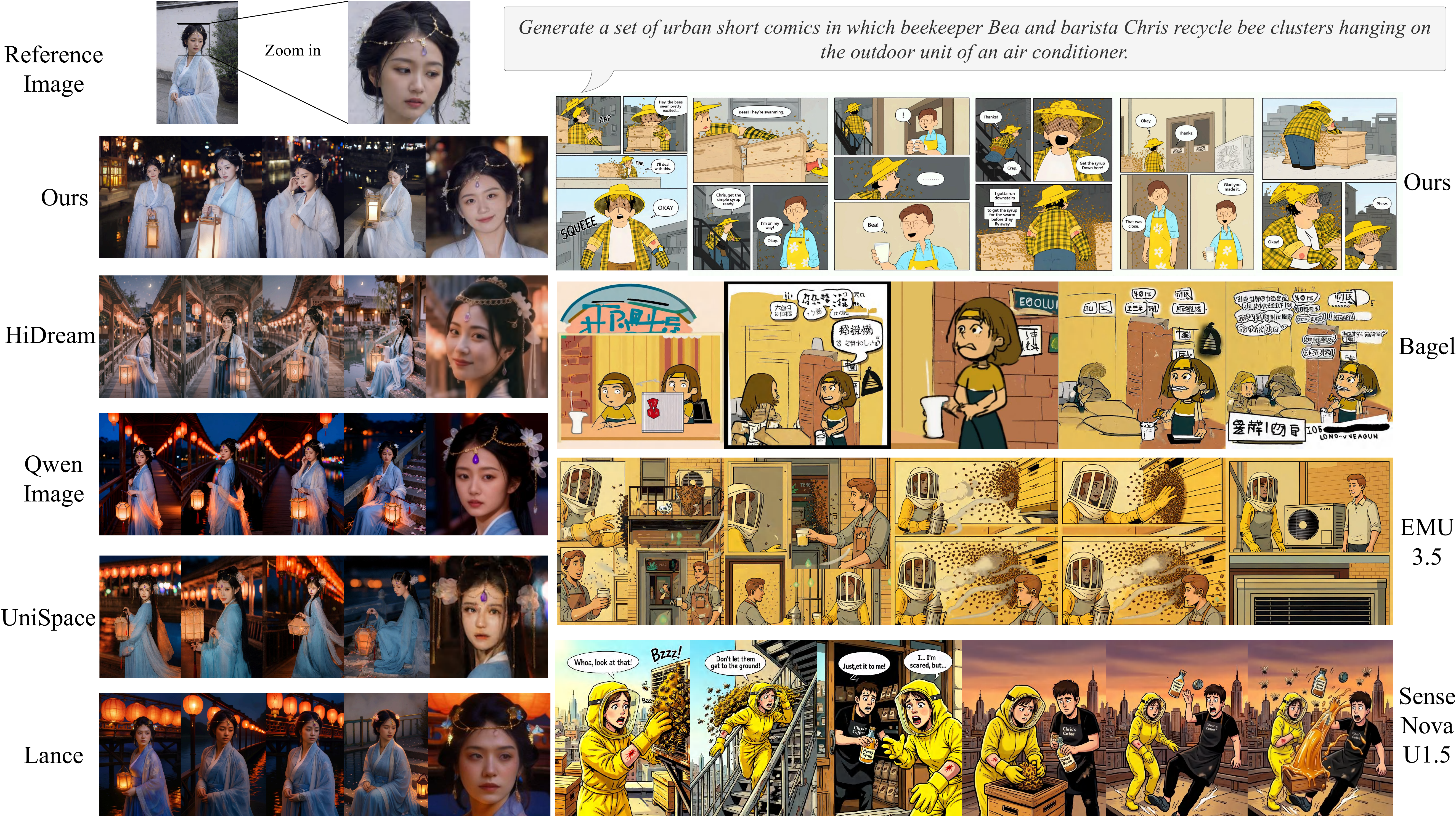}
}
\caption{\textbf{High-fidelity long-form generation.} Left: image-conditioned generation with identity preservation across diverse views. Right: long-form interleaved generation with consistent characters and visual states over an extended trajectory.}
\label{fig:qualitative}
\end{figure*}

Figure~\ref{fig:qualitative} further highlights the capability gap between simply supporting interleaved generation and sustaining high-quality generation over long trajectories. In image-conditioned generation, Ours preserves fine-grained facial appearance while producing diverse, photorealistic compositions rather than repeatedly copying a small set of poses or layouts. These results demonstrate that our setting goes beyond short or visually simplified interleaved generation and preserves strong native image-generation and multimodal-planning capabilities over extended trajectories.

\subsection{Scaling Native Interleaved Generation to Longer Horizons}
\label{sec:scaling}

\begin{figure*}[t]
\centering
\includegraphics[width=\textwidth]{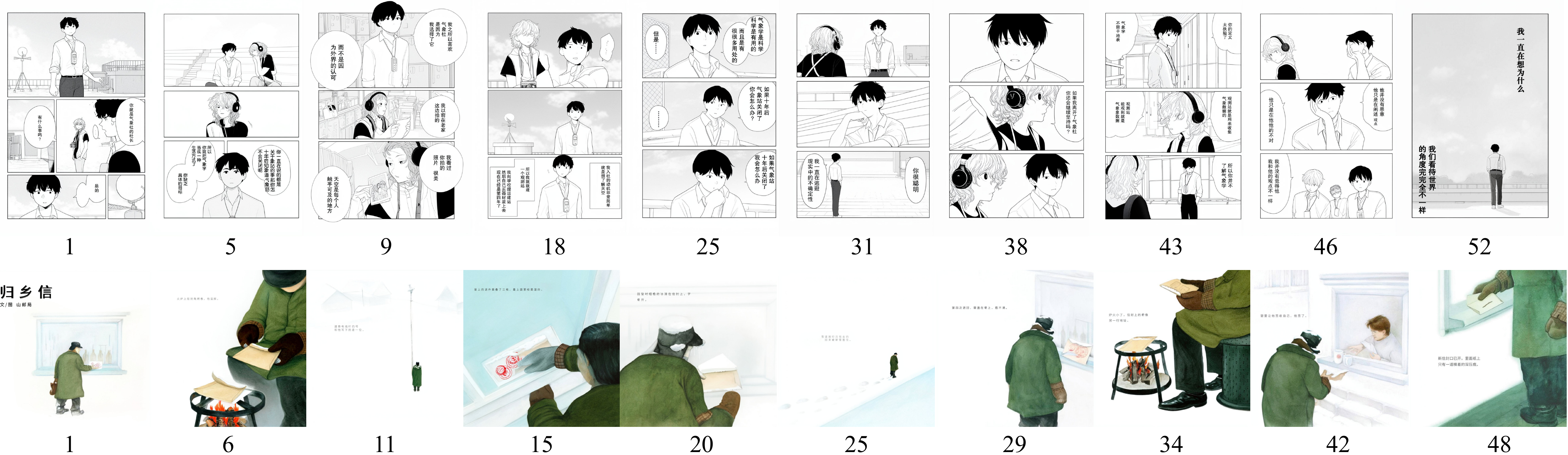}
\caption{\textbf{Native interleaved generation over extended trajectories.}
Representative frames from long autoregressive generations across diverse visual domains, including 50+ image trajectories.}
\label{fig:long_horizon}
\end{figure*}

LongMIG-Bench focuses on a practical generation horizon that enables broad comparison across existing systems. We further examine whether the native interleaved capability of our model can persist substantially beyond this regime. As shown in Figure~\ref{fig:long_horizon}, the model can sustain autoregressive text-image generation over dozens of image-generation steps, including trajectories extending beyond 50 images. Across widely separated generations, recurring characters and visual states remain recognizable, while the model continues to introduce new compositions and content rather than simply reproducing earlier frames. These examples highlight an important property of our formulation: extending the generation horizon does not require truncating the multimodal trajectory. The complete semantic history remains accessible to the understanding pathway, while scene-aware visual memory controls the dense historical representations directly used for image synthesis.

\subsection{Ablation Study}
\label{sec:ablation}

We conduct stage-specific controlled comparisons of scene-aware retrieval, persistent visual references, and self-rollout adaptation, as shown in Table~\ref{tab:component_ablation}. We use the 24-image \textsc{Long+Ext} stress test for these ablations, as the longer horizon more clearly exposes differences in memory behavior; LR metrics focus on scene revisits separated by at least eight generation steps.

\paragraph{Scene-aware retrieval.}
The first block evaluates how scene information affects visual-memory retrieval. Predicted routing performs comparably to retrieval using annotated scene labels, with small differences across scene and long-range metrics. In contrast, corrupting $30\%$ of the scene signal reduces Scene from $3.26$ to $2.81$ and LR Scene from $3.10$ to $2.76$. Replacing scene-aware retrieval with fixed recent windows causes similar degradation, worsening when the window is reduced from four to two images. These results show that long-range generation benefits from retrieving semantically relevant historical states rather than relying on recent visual context.

\begin{table*}[t]
\centering
\caption{\textbf{Controlled analysis of visual-memory components.}
Ablations of scene-aware retrieval, persistent visual references, and self-rollout adaptation on the 24-image \textsc{Long+Ext} stress test.}
\label{tab:component_ablation}
\small
\setlength{\tabcolsep}{3.8pt}
\resizebox{1\textwidth}{!}{
\begin{tabular}{llcccccc}
\toprule
Variant & Setting & Identity $\uparrow$ & Scene $\uparrow$ & Gap-D $\uparrow$ & LR Scene $\uparrow$ & LR Gap-D $\uparrow$ & Visual Fidelity $\uparrow$ \\
\midrule

\multicolumn{8}{l}{\textit{Scene-aware retrieval}} \\
Predicted Routing & Predicted Scene & \textbf{3.36} & 3.26 & 0.164 & 3.10 & 0.119 & \textbf{3.15} \\
GT Scene & Annotated Scene & 3.21 & \textbf{3.32} & \textbf{0.173} & \textbf{3.15} & \textbf{0.121} & 3.03 \\
Noisy Scene & 30\% Corrupted & 3.30 & 2.81 & 0.145 & 2.76 & 0.099 & 3.14 \\
Recent-4 & Scene-Agnostic & 3.15 & 2.85 & 0.147 & 2.79 & 0.092 & 2.97 \\
Recent-2 & Scene-Agnostic & 3.05 & 2.58 & 0.114 & 2.47 & 0.071 & 2.89 \\

\midrule
\multicolumn{8}{l}{\textit{Persistent visual references}} \\
With Reference & Persistent Reference & \textbf{3.36} & \textbf{3.26} & \textbf{0.164} & \textbf{3.10} & \textbf{0.119} & \textbf{3.15} \\
w/o Reference & No Reference & 3.18 & 2.77 & 0.048 & 2.03 & 0.005 & 3.13 \\

\midrule
\multicolumn{8}{l}{\textit{Self-rollout adaptation}} \\
Base & No Additional FT & 2.89 & 2.57 & 0.193 & 2.51 & 0.157 & 2.57 \\
GT-FT & GT History & 3.23 & 2.71 & 0.191 & 2.52 & 0.161 & 3.35 \\
Self-Rollout & Generated History & \textbf{3.59} & \textbf{3.25} & \textbf{0.194} & \textbf{3.12} & \textbf{0.174} & \textbf{3.38} \\

\bottomrule
\end{tabular}}
\end{table*}

\paragraph{Persistent visual references.}
The second block studies the role of persistent visual references. Removing reference-based training reduces Identity from $3.36$ to $3.18$, confirming that persistent references help stabilize recurring character appearance over long trajectories. Scene consistency also degrades, with Scene decreasing from $3.26$ to $2.77$ and LR Scene from $3.10$ to $2.03$; Gap-D and LR Gap-D similarly drop from $0.164$ to $0.048$ and from $0.119$ to $0.005$. In contrast, Visual Fidelity remains nearly unchanged ($3.15$ vs.\ $3.13$). These results suggest that persistent references primarily provide stable identity anchors while also benefiting long-range visual consistency.

\paragraph{Self-rollout adaptation.}
The final block studies the effect of training on self-generated histories. Additional fine-tuning with ground-truth histories substantially improves Visual Fidelity over the base model ($3.35$ vs.\ $2.57$), but provides only limited gains in long-range consistency. Under the same additional training data, Self-Rollout further improves Identity from $3.23$ to $3.59$, Scene from $2.71$ to $3.25$, and LR Scene from $2.52$ to $3.12$, while maintaining comparable Visual Fidelity ($3.38$). Gap-D and LR Gap-D remain relatively stable, indicating that the primary benefit of self-rollout lies in maintaining coherent identities and scenes when the model must continue from its own imperfect generations.

\begin{figure*}[t]
\centering
\includegraphics[width=\textwidth]{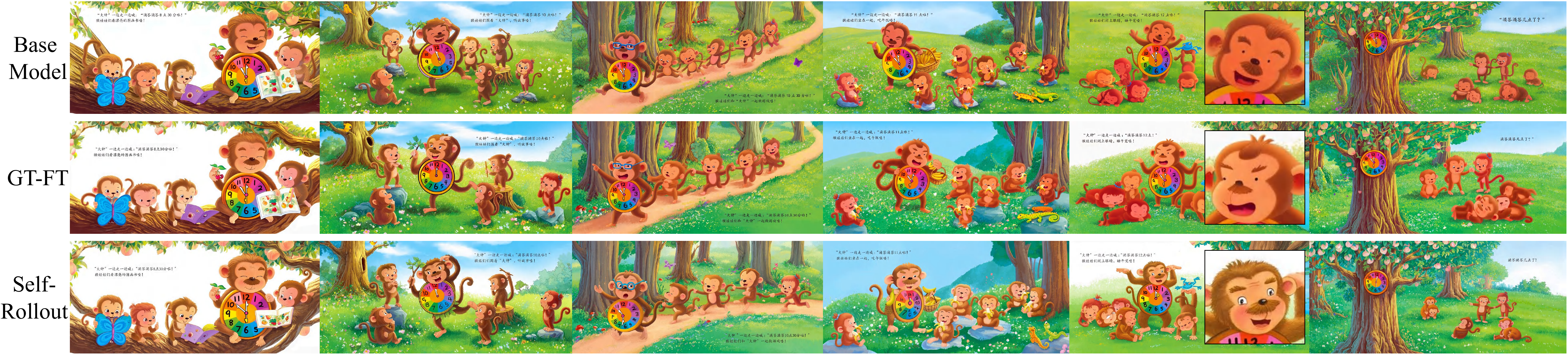}
\caption{\textbf{Effect of self-rollout adaptation.}
Base and GT-FT exhibit progressive reddish color drift and loss of fine-grained facial details, whereas Self-Rollout better preserves color appearance and local visual structure throughout the trajectory.}
\label{fig:self_rollout}
\end{figure*}

Figure~\ref{fig:self_rollout} qualitatively illustrates this effect. Base and GT-FT exhibit color drift and loss of fine-grained facial details in later generations, whereas Self-Rollout better preserves color stability and visual structure. Together with the quantitative results, this suggests that exposure to self-generated histories is important for sustaining high-quality generation over long autoregressive trajectories.

\section{Conclusion}
\label{sec:conclusion}

We develop a large UMM natively trained for interleaved text-image generation and study how to extend its strong image synthesis and multimodal planning capabilities to long autoregressive trajectories. We introduce a visual-memory framework that combines persistent visual references, scene-aware retrieval, and self-rollout adaptation while preserving the complete semantic history for reasoning. Experiments on LongMIG-Bench demonstrate high-fidelity long-horizon generation with strong character and scene consistency, and controlled studies validate the complementary roles of the proposed components. Beyond the benchmark horizon, our model further sustains native interleaved generation over extended trajectories spanning dozens of images. Overall, our results show that explicit visual-memory management is important for scaling native interleaved generation to substantially longer horizons.

\bibliography{iclr2027_conference}
\bibliographystyle{iclr2027_conference}
\clearpage

\appendix
\begingroup
\setlength{\parskip}{3pt}
\linespread{1.0}\selectfont

\section{Additional Qualitative Results}
\label{app:qualitative}

We provide additional qualitative results to complement the quantitative evaluation in the main paper. We first compare complete long-form trajectories with representative baselines on multi-step visual process generation, then demonstrate generation across diverse visual domains, and finally present a complete native interleaved planning-and-generation trajectory produced by our model.

\subsection{Long-Form Generation Comparisons}
\label{app:qualitative_comparison}

Figures~\ref{fig:qualitative_eggtart} and~\ref{fig:qualitative_pork} compare complete generation trajectories on two multi-step cooking tasks. Unlike isolated image generation, these examples require a model to represent distinct intermediate states and maintain meaningful progression throughout the sequence.

In the egg-tart example, several methods produce plausible individual images but repeatedly return to similar compositions or already completed food states, resulting in limited progression across the trajectory. Some interleaved models also reuse nearly identical visual layouts across multiple steps. Our model instead depicts a broader sequence of preparation, filling, baking, inspection, and serving stages, with substantial changes in viewpoint and composition across the trajectory.

The braised-pork example exhibits a similar distinction. The requested sequence requires the ingredients to undergo successive transformations rather than repeatedly showing the same intermediate or final dish. Our model progresses from ingredient preparation through cooking and thickening to the completed dish while maintaining a coherent illustrative style. Together, these examples show that strong individual images do not necessarily imply coherent multi-step visual progression over a long generation trajectory.

\begin{figure*}[t]
\centering
\includegraphics[width=\textwidth]{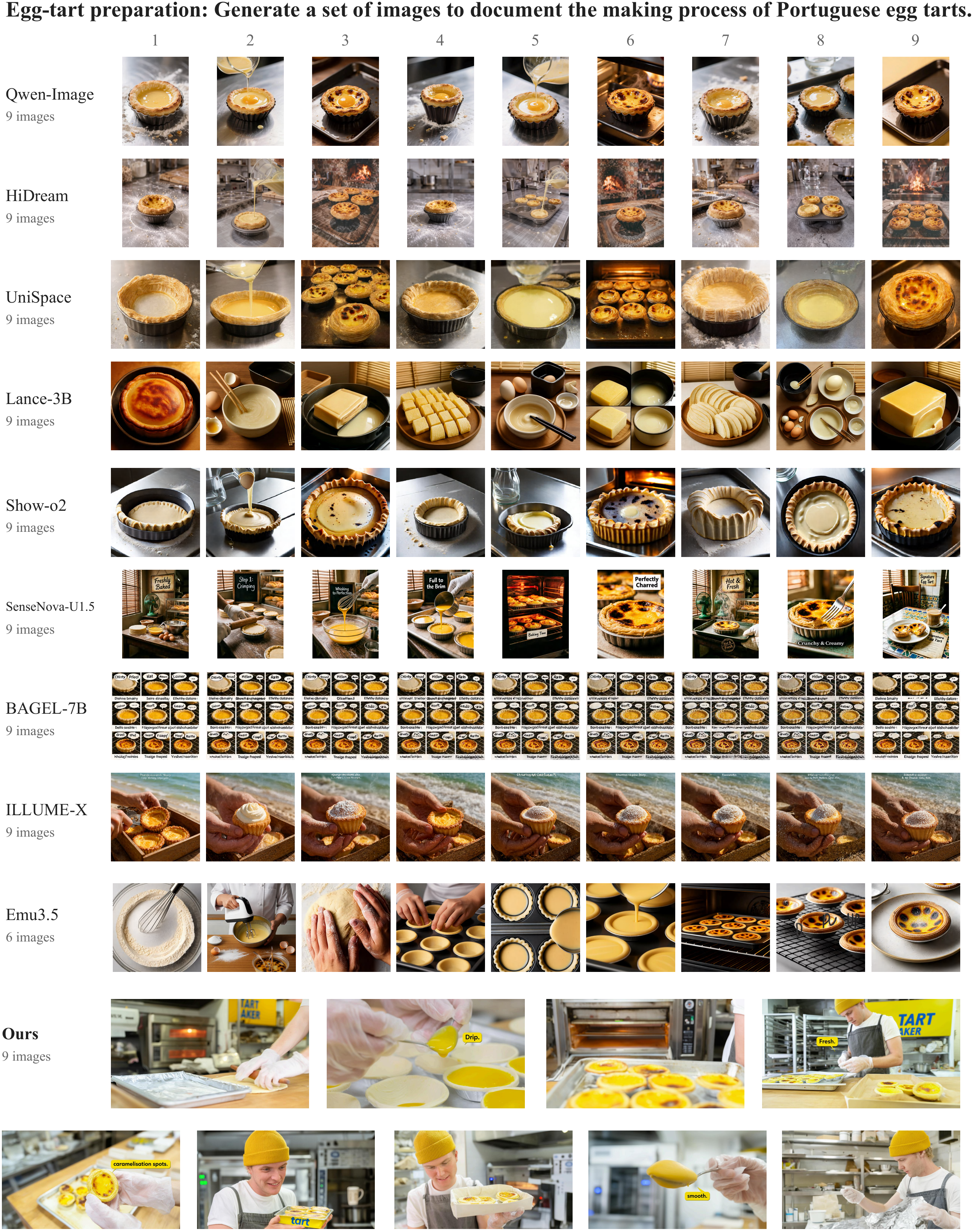}
\caption{\textbf{Long-form visual process generation for Portuguese egg tarts.}
All methods are prompted to generate a sequence documenting the preparation process. The comparison highlights differences in step progression, visual diversity, repetition, and consistency across the complete trajectory.}
\label{fig:qualitative_eggtart}
\end{figure*}

\begin{figure*}[t]
\centering
\includegraphics[width=\textwidth]{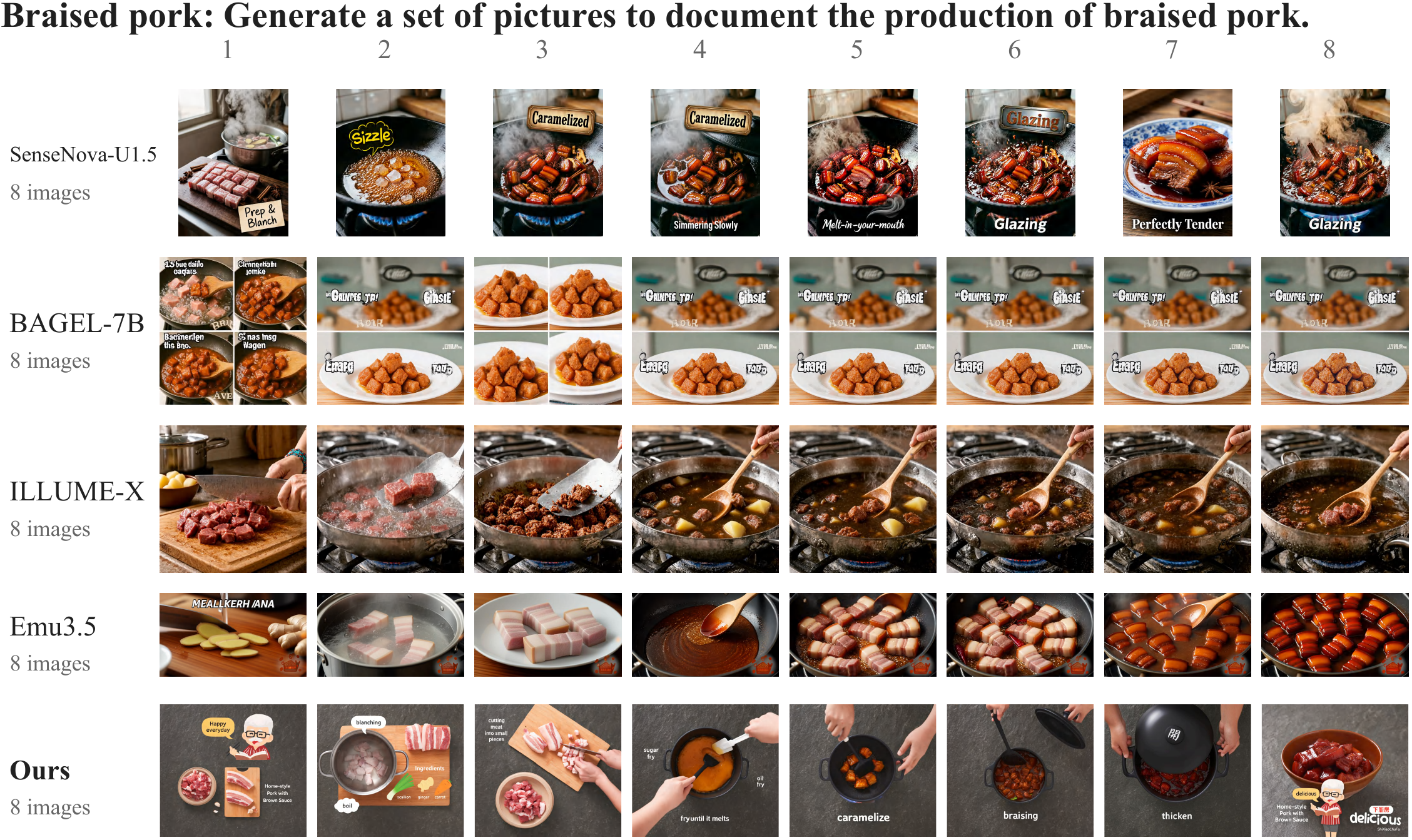}
\caption{\textbf{Long-form visual process generation for braised pork.}
Each method generates a multi-image sequence describing the cooking process. The examples illustrate how models differ in representing successive visual states rather than repeatedly depicting similar intermediate or final results.}
\label{fig:qualitative_pork}
\end{figure*}

\subsection{Generation Across Diverse Visual Domains}
\label{app:qualitative_domains}

Figure~\ref{fig:qualitative_domains} presents additional long-form trajectories across a diverse range of visual domains, including cinematic storytelling, cooking photography, minimalist typography, architectural visualization, illustrated recipes, graphic identity design, product design, and pet photography. These examples demonstrate that long-horizon interleaved generation is not limited to character-centric stories or a single visual style.

Different domains place different demands on long-form generation. Cinematic and photographic sequences require recurring subjects and environments to remain recognizable across changing viewpoints and compositions. Typography and branding tasks require consistent graphic language, layout, and text rendering across multiple outputs. Multi-step visual tasks require meaningful semantic progression, while image-conditioned product and pet examples require a user-provided visual identity to be preserved across substantially different contexts. The same unified model supports these different generation modes within a common interleaved framework.

\begin{figure*}[p]
\centering
\includegraphics[width=\textwidth,height=0.82\textheight,keepaspectratio]{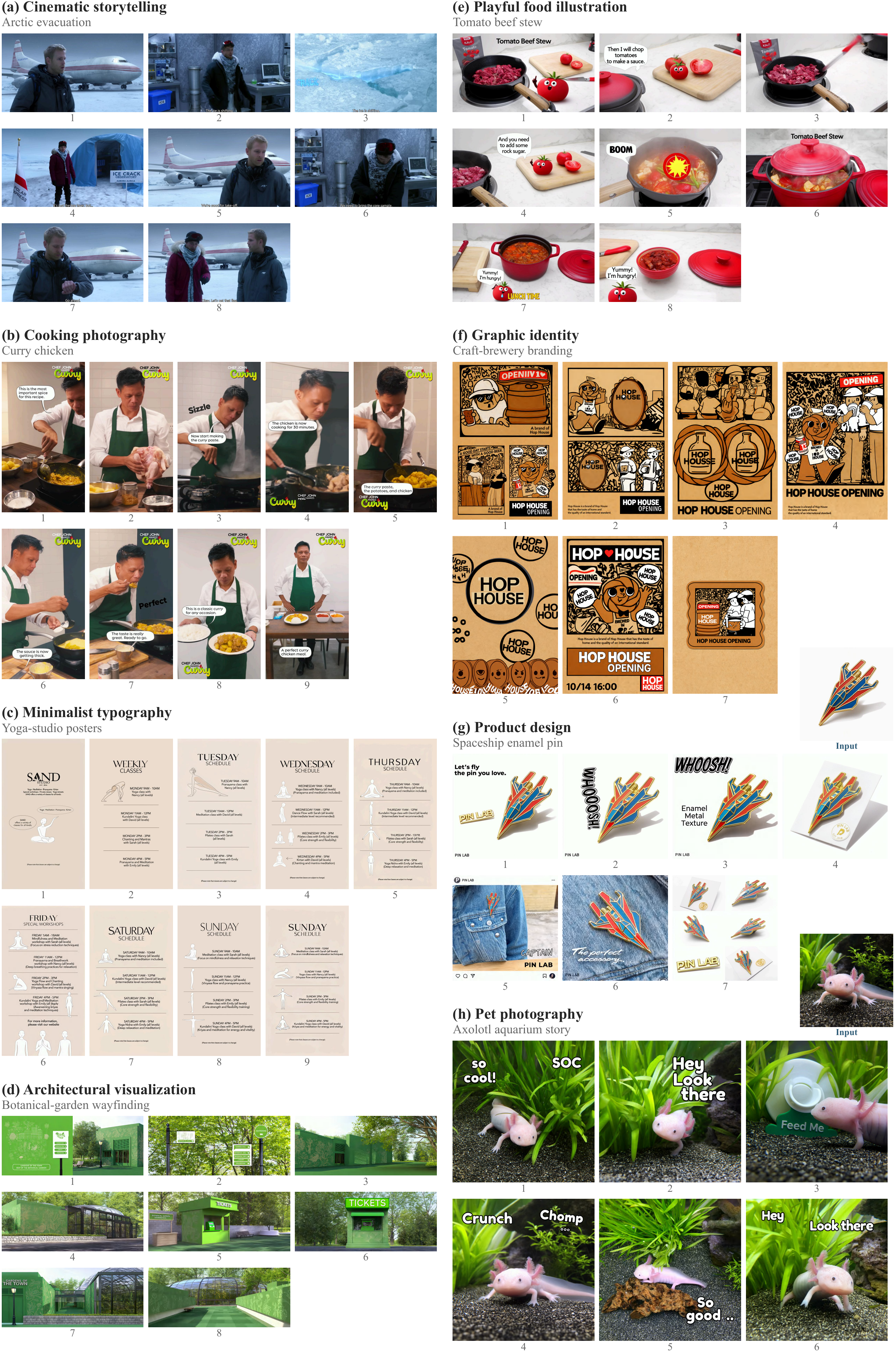}
\caption{\textbf{Long-form generation across diverse visual domains.}
Representative trajectories cover (a) cinematic storytelling, (b) cooking photography, (c) minimalist typography, (d) architectural visualization, (e) playful food illustration, (f) graphic identity design, (g) image-conditioned product design, and (h) image-conditioned pet photography. Input reference images for image-conditioned examples are shown separately.}
\label{fig:qualitative_domains}
\end{figure*}

\subsection{Complete Multimodal Planning and Generation Trajectory}
\label{app:qualitative_trajectory}

Figure~\ref{fig:complete_trajectory} shows a complete native interleaved trajectory, including both intermediate textual planning and generated visual states. Given a request to create an eight-page nutrition presentation, the model first forms an overall content and visual plan, generates a reusable visual reference for the recurring character, and organizes the requested content into a sequence of slide-level scenes. It then alternates between textual planning and image generation until the complete presentation is produced.

The example illustrates how planning and visual synthesis occur within the same autoregressive multimodal trajectory. Each generated image is inserted back into the context and can influence subsequent planning and generation. Across the sequence, the model reuses the recurring character and overall graphic language while changing the slide content, layout, food imagery, and instructional message according to the evolving plan.

This differs from a pipeline in which an external planner first decomposes the task into independent prompts and a separate image generator renders each output in isolation. Here, intermediate text and generated visual states remain part of the same evolving multimodal context throughout generation.

\begin{figure*}[p]
\centering
\includegraphics[width=\textwidth,height=0.82\textheight,keepaspectratio]{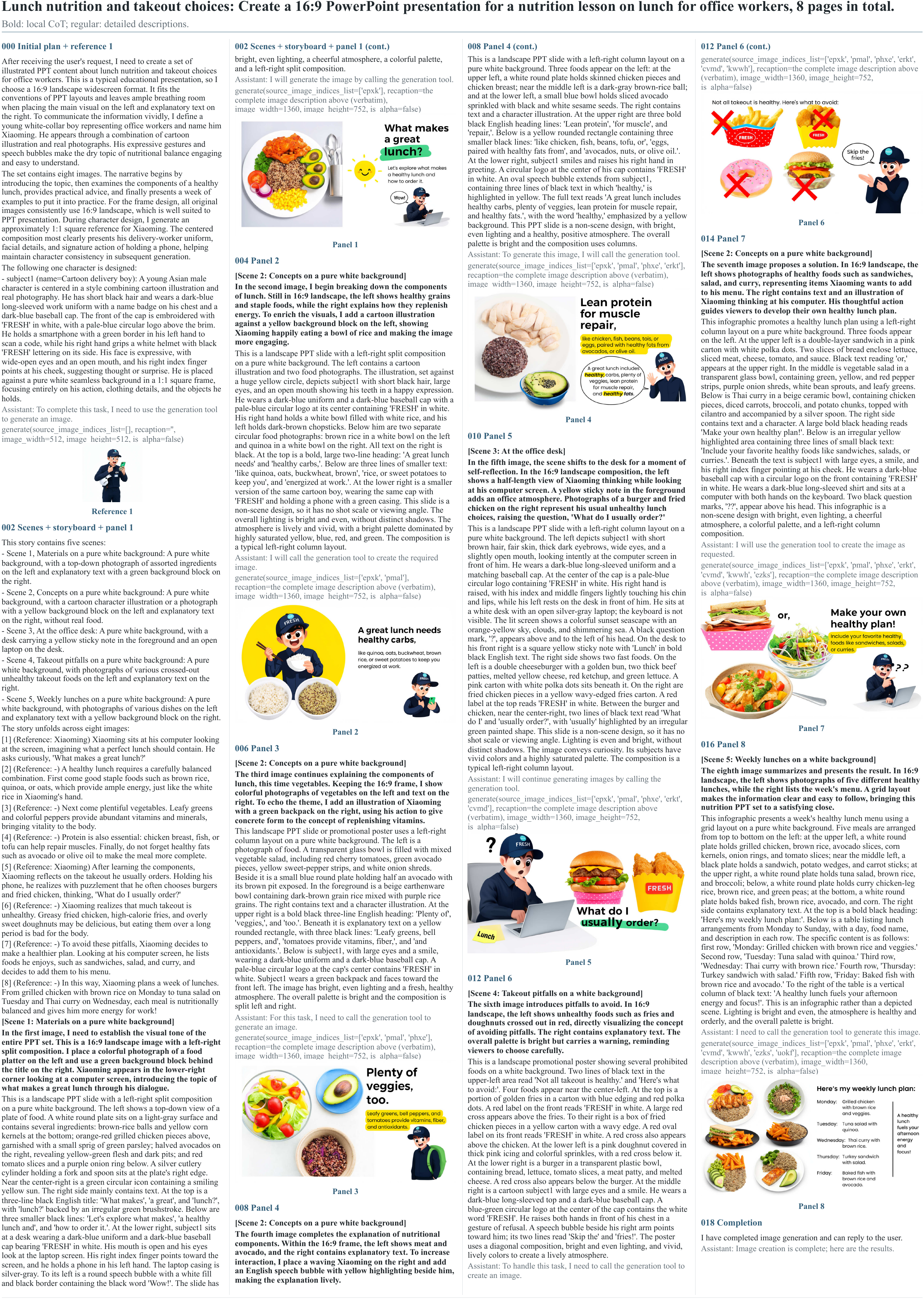}
\caption{\textbf{Complete native interleaved planning-and-generation trajectory.}
Given a request for an eight-page nutrition presentation, the model first forms an overall plan and generates a reusable visual reference, then alternates between textual planning and image generation until completing the requested sequence. Generated images are inserted back into the multimodal context and remain available to subsequent generation steps.}
\label{fig:complete_trajectory}
\end{figure*}

\par\endgroup
\clearpage
\section{Training Data Construction}
\label{app:data}

\subsection{Data Pipeline and Scale}
\label{app:data_pipeline}

The overall training pipeline uses progressively curated multimodal data, moving from broad multimodal coverage toward increasingly specialized long-horizon supervision. We begin with a large-scale candidate pool on the order of $10^8$ examples, covering natural images, image--text pairs, image-conditioned generation data, multi-image sequences, comics, illustrated stories, and design-oriented content. The candidate pool is progressively processed through quality filtering, deduplication, resolution and file-integrity checks, semantic consistency filtering, and task-specific parsing.

The resulting data are organized into several stages with increasingly specialized supervision. Pre-training (PT) uses data on the order of $10^6$ examples to learn general multimodal understanding and generation. Supervised fine-tuning (SFT) further uses on the order of $10^5$ curated examples spanning single-image generation, image-conditioned generation, multi-image generation, and native text--image interleaving. From this mixture, we construct a dedicated long-horizon corpus containing recurring entities, explicit scene transitions, and long-range scene revisits. Finally, self-rollout adaptation uses a smaller set of trajectories augmented with the model's own generated histories. 

\subsection{Long-Horizon Corpus Composition}
\label{app:data_composition}

The long-horizon post-training corpus contains $21{,}162$ interleaved stories with $197{,}653$ generated images, including $53{,}499$ persistent visual reference images and $144{,}154$ story panels. Reference images precede the corresponding story panels and provide reusable visual states for recurring entities. Each story contains between $2$ and $24$ story panels and between $1$ and $10$ visual references, with a mode of two.

The corpus is constructed from several types of sequential visual data, including multi-shot storyboards, illustrated children's books, social-media comics, long-form comics with recurring environments, and design-oriented sequences. The source distribution is summarized in Table~\ref{tab:data_composition}.

\begin{table}[t]
\centering
\caption{\textbf{Composition of the long-horizon post-training corpus.}}
\label{tab:data_composition}
\small
\setlength{\tabcolsep}{6pt}
\begin{tabular}{lc}
\toprule
Data type & Fraction \\
\midrule
Multi-shot storyboards & 45.3\% \\
Illustrated books with visual references & 18.5\% \\
Illustrated books without visual references & 7.9\% \\
Social-media comics & 14.5\% \\
Long comics with background continuity & 9.1\% \\
Brand and design sequences & 4.7\% \\
\bottomrule
\end{tabular}
\end{table}

For sequential sources, we additionally parse recurring entities and physical scenes and convert the retained examples into a unified long-form interleaved format. Samples with missing images, ambiguous or unresolvable scene structure, duplicate content, or trajectories exceeding the training budget are removed during this process.

\subsection{Interleaved Sequence Construction}
\label{app:data_format}

Each long-horizon example is represented as a multi-turn multimodal conversation in which textual planning and image generation are interleaved within a single trajectory. Before generating an image, the assistant may produce textual reasoning describing the intended content and its relation to the preceding trajectory, followed by an image-generation call. The generated image is then inserted back into the sequence and becomes available to subsequent text and image generation.

A trajectory can be written schematically as
\begin{equation}
x
\rightarrow
r_1, I_1
\rightarrow
r_2, I_2
\rightarrow \cdots \rightarrow
r_N, I_N,
\end{equation}
where $x$ is the initial instruction, $r_k$ denotes the textual planning associated with generation step $k$, and $I_k$ is the corresponding generated image.

Every generated image participates in the sequence in two complementary forms. A noised latent is used as the flow-matching target for the current generation step, while a clean visual representation is retained as historical conditioning for subsequent generations. Thus, later generations can directly depend on visual states produced earlier in the same trajectory rather than treating each image as an independent training example.

Later image-generation calls also retain references to previously generated images, allowing the training sequence to preserve the complete multimodal trajectory and to supervise dependencies spanning multiple generation steps.

\subsection{Persistent References and Scene Annotations}
\label{app:data_annotations}

Long-form trajectories contain two additional types of structure used by the visual-memory mechanism.

First, auxiliary visual-reference images are identified as \emph{persistent visual references}. These images are generated before the corresponding story panels when recurring visual entities need to be established and are trained with the same native image-generation objective as ordinary outputs. Unlike story panels, they are treated as persistent generative memory and remain accessible throughout the subsequent trajectory.

Second, each story panel is assigned a discrete scene state according to its underlying physical environment. Images depicting the same physical location retain the same scene state despite changes in camera viewpoint, composition, character pose, or local action. Persistent reference images are excluded from scene matching and are handled independently by the persistent-memory rule.

These annotations are used to supervise scene-state prediction during long-horizon post-training. At inference time, no ground-truth scene annotation is provided: the model predicts the upcoming scene state from its multimodal history, and this prediction is subsequently used for scene-aware visual-memory retrieval.

For the no-reference ablation, we rebuild the same training episodes after removing all persistent-reference generation calls while preserving the remaining story panels and their ordering.

\subsection{Self-Rollout Data}
\label{app:rollout_data}

Self-rollout uses a parallel version of the long-horizon corpus in which selected historical conditioning images are replaced by outputs generated by the model itself. The rollout images are produced offline and written back into the corresponding training trajectories; no autoregressive image sampling is performed inside the optimization loop.

The two rollout stages contain $5{,}721$ and $7{,}522$ stories, respectively, for a total of $13{,}243$ rollout trajectories. The second stage places greater emphasis on portrait and social-media sequences, where identity drift and visual error accumulation are particularly common.

For a selected rollout position, only the clean historical conditioning copy is replaced by the model-generated image. The noised image used as the flow-matching supervision target remains the original ground-truth image. Persistent visual references are never replaced, and the first two non-reference story images are kept clean to provide a reliable initial visual state. This construction exposes the model to realistic inference-time histories without directly treating its own previous errors as training targets.

\section{LongMIG-Bench Construction}
\label{app:benchmark}

\subsection{Benchmark Composition}
\label{app:benchmark_composition}

LongMIG-Bench contains $100$ long-form story specifications, evenly divided between text-to-multi-image (\textsc{T2MI}) and image-to-multi-image (\textsc{I2MI}) generation. The \textsc{T2MI} track contains $50$ stories with $462$ specified shots, while the \textsc{I2MI} track contains $50$ stories with $460$ shots and additionally provides a user reference photograph for each story. Stories contain $9.2$ shots on average and cover illustrated children's books, comics, storyboards, portrait series, pet and product photography, recipes, poster design, sticker sets, and landscape compositions.

To avoid evaluating only a narrow visual regime, each story specifies its own visual style and target aspect ratio. The benchmark contains $50$ distinct styles in each track and seven aspect ratios ranging from $9{:}16$ to $16{:}9$. Every specification is available in both English and Chinese.

\begin{table}[t]
\centering
\caption{\textbf{Composition of LongMIG-Bench.}}
\label{tab:benchmark_composition}
\small
\setlength{\tabcolsep}{6pt}
\begin{tabular}{lcc}
\toprule
& \textsc{T2MI} & \textsc{I2MI} \\
\midrule
Stories & 50 & 50 \\
Specified shots & 462 & 460 \\
Avg. shots / story & \multicolumn{2}{c}{9.2} \\
Min. revisited shots / story & \multicolumn{2}{c}{$\geq 4$} \\
Avg. revisited shots & 6.16 & 6.04 \\
Distinct styles & 50 & 50 \\
User reference image & No & Yes \\
\bottomrule
\end{tabular}
\end{table}

\subsection{Structured Story Specifications}
\label{app:benchmark_structure}

Each story is represented as structured data rather than free-form prose. A specification contains a cast list with per-character visual attributes, a location table, an ordered shot sequence, and shot-level annotations describing the corresponding scene, whether the scene revisits an earlier location, and which entities should persist from previous shots. This structure provides explicit supervision for evaluating identity preservation, scene recurrence, and long-range visual consistency.

A simplified specification can be represented as
\begin{equation}
\mathcal{S}=
\{\mathcal{C},\mathcal{L},(q_1,s_1,r_1,e_1),\ldots,(q_N,s_N,r_N,e_N)\},
\end{equation}
where $\mathcal{C}$ and $\mathcal{L}$ denote the character and location definitions, $q_k$ is the specification for shot $k$, $s_k$ is its scene identifier, $r_k$ indicates whether the scene is a revisit, and $e_k$ specifies entities carried over from the preceding trajectory.

\subsection{Long-Range Revisit Design}
\label{app:benchmark_revisit}

Long-range scene revisits are introduced deliberately rather than occurring incidentally. Every story contains at least four revisited shots, with an average of $6.16$ revisits in \textsc{T2MI} and $6.04$ in \textsc{I2MI}. In these trajectories, scenes established earlier in the story reappear after multiple intervening shots and scene transitions.

This design is important for evaluating visual memory. If scenes occur only once, or if every revisit falls within a short recency window, scene-aware retrieval becomes difficult to distinguish from simply retaining the most recent images. LongMIG-Bench instead contains repeated long-range revisits for which the relevant visual state may lie far outside the recent context, directly testing whether a model can recover previously established visual information.

\subsection{Benchmark Examples}
\label{app:benchmark_examples}

\begin{figure*}[t]
\centering
\includegraphics[width=\textwidth]{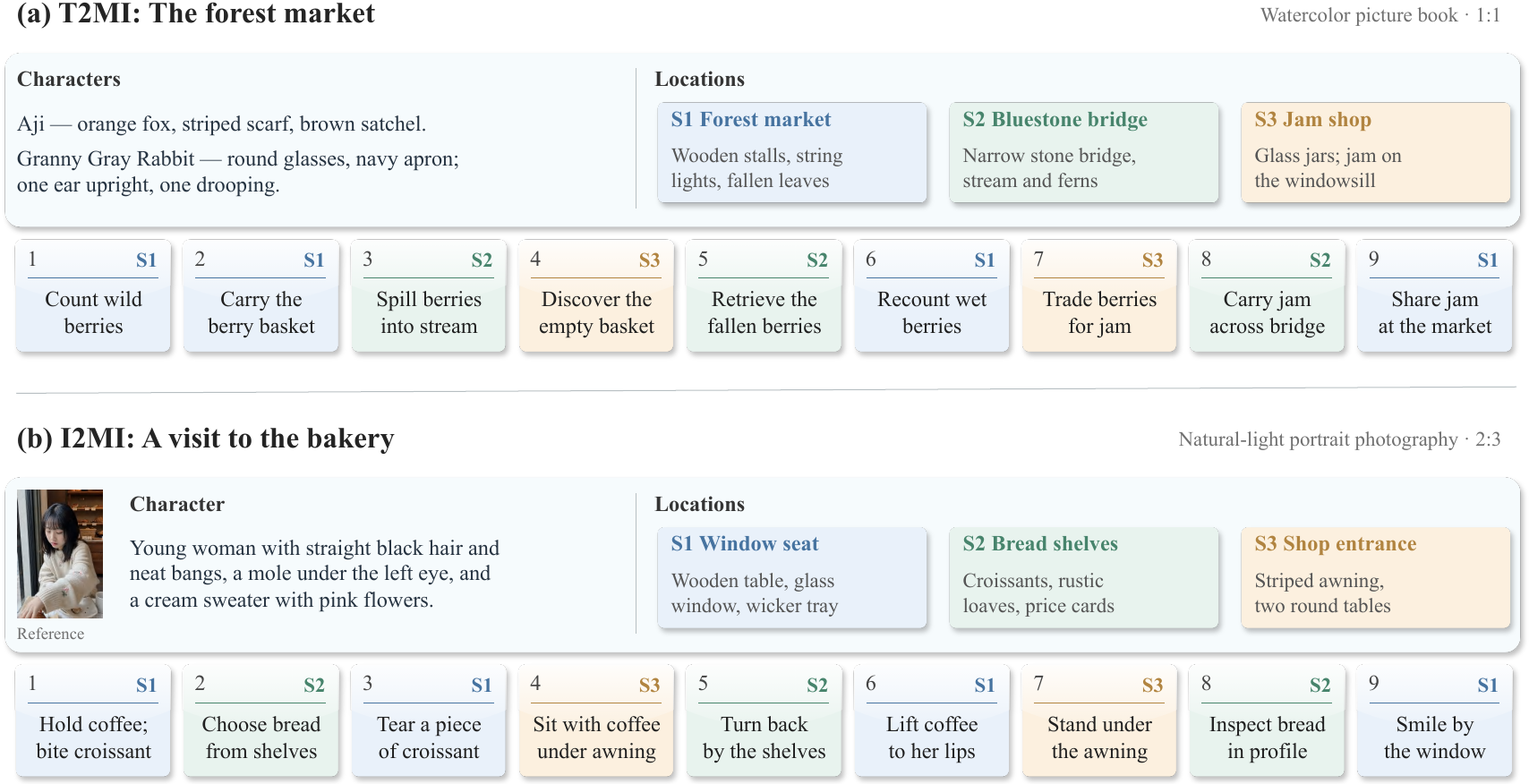}
\caption{\textbf{Examples from LongMIG-Bench.} Each story contains structured character and location definitions together with an ordered sequence of shots. Colors indicate recurring scenes, and arrows highlight long-range revisits after intervening scene transitions. The \textsc{I2MI} track additionally provides a user reference image.}
\label{fig:benchmark_examples}
\end{figure*}

Figure~\ref{fig:benchmark_examples} visualizes representative specifications from both benchmark tracks. In addition to the natural-language story instruction, we show the underlying character definitions, scene structure, shot sequence, and long-range revisit relationships. These structured annotations are used only for benchmark construction and evaluation; models receive the corresponding story-level generation instruction rather than ground-truth scene annotations during open-ended generation.

\subsection{Long+Ext Long-Horizon Stress Test}
\label{app:long_ext}

In addition to LongMIG-Bench, we construct \textsc{Long+Ext}, a locked-text stress-test suite for controlled long-horizon evaluation. Every \textsc{Long+Ext} story contains exactly $24$ predefined image slots with fixed shot descriptions and scene assignments.

\textsc{Long+Ext} contains $38$ stories in total. Each story contains two recurring characters and three recurring scenes that are repeatedly revisited throughout the $24$-image trajectory. The scene-transition patterns are fixed during benchmark construction to ensure frequent non-adjacent revisits rather than contiguous scene blocks.

All methods are evaluated on the same shot sequence and scene structure. Native interleaved models receive the complete story-level instruction, whereas independent image generators receive corresponding self-contained per-shot prompts.

Long-range (LR) metrics evaluate only revisits occurring at least eight generation steps after the \emph{first establishing view} of the corresponding scene:
\begin{equation}
k-e_s \geq 8,
\end{equation}
where $e_s$ is the first occurrence of scene $s$. This setting specifically tests whether a model can recover distant visual states beyond a short recent-context window. \textsc{Long+Ext} is used primarily for the controlled component ablations in Table~\ref{tab:component_ablation}.

\section{Evaluation Metrics}
\label{app:evaluation}

\paragraph{Character Identity.}
\textbf{Character Identity} measures whether each named recurring character remains visually recognizable across the frames in which it appears. The evaluator considers identity-defining attributes such as facial appearance, body shape, hair or fur characteristics, clothing, and other persistent details, while allowing pose, expression, viewpoint, and illumination to vary. Repeating nearly identical images can trivially inflate identity consistency, so this metric is interpreted jointly with Scene Memory and Visual Memory.

\paragraph{Scene Memory.}
\textbf{Scene Memory} measures whether a previously established physical environment is correctly recovered after the trajectory leaves and later revisits it. The evaluator considers persistent architectural structure, spatial layout, color palette, signage, and other location-specific visual cues, while also requiring distinct environments to remain distinguishable. To prevent degenerate background copying, a trajectory that effectively reuses the same background for all scenes receives a Scene score of at most $2$.

For long-range analysis on \textsc{Long+Ext}, \textbf{LR Scene} restricts evaluation to revisits occurring at least eight generation steps after the first establishing view of the corresponding scene.

\paragraph{Visual Memory.}
\textbf{Visual Memory} measures whether later generations preserve visual decisions introduced by earlier generated frames rather than reconstructing each image solely from the textual specification. Such information may include the position of a facial mark, the layout of a storefront sign, a garment pattern, or other instance-specific details that emerge during generation. Preserving identity directly from a user-provided reference image does not by itself constitute visual memory. Conversely, trivially repeating the same frame throughout the trajectory is assigned the minimum score.

\paragraph{Visual Fidelity and Aesthetic Quality.}
\textbf{Visual Fidelity} evaluates the technical quality of generated images within the visual style requested by the story, including anatomy, composition, rendering artifacts, and malformed text. Stylized, grayscale, comic, watercolor, or other non-photorealistic outputs are evaluated relative to their requested style rather than against a photorealistic standard.

\textbf{Aesthetic Quality} evaluates overall visual craftsmanship and appeal within the declared style, including composition, graphic coherence, visual polish, and style-appropriate presentation. Photorealism is not rewarded unless explicitly requested. Severe noise, malformed anatomy, corrupted text, inappropriate borders, or repetitive use of a single visual layout reduce the score.

\paragraph{Text Rendering.}
\textbf{Text Rendering} evaluates the visual quality, legibility, and semantic faithfulness of story-relevant text rendered within generated images, including storefront signs, posters, menus, titles, dialogue, and other scene-integrated textual elements. The evaluator considers whether characters and words are clearly formed, readable, visually coherent with the surrounding content, and free from common rendering artifacts such as malformed glyphs, distorted strokes, or inconsistent typography. This metric reflects the model's ability to preserve strong text-generation capability within long-horizon interleaved image sequences.

\paragraph{SceneGap.}
We additionally measure scene separability using
\begin{equation}
\mathrm{SceneGap}
=
\bar{s}_{\mathrm{same}}
-
\bar{s}_{\mathrm{cross}},
\end{equation}
where $\bar{s}_{\mathrm{same}}$ is the average visual similarity between images associated with the same scene and $\bar{s}_{\mathrm{cross}}$ is the average similarity between images from different scenes. We compute SceneGap using SigLIP2 and DINOv3 representations, denoted \textbf{Gap-S} and \textbf{Gap-D}, respectively. Positive values indicate that recurring scenes are visually closer to themselves than to other environments.

\subsection{Trajectory-Level Judge}
\label{app:judge}

We use GPT-6 as the trajectory-level evaluator for Character Identity, Scene Memory, Visual Memory, Visual Fidelity, Aesthetic Quality, and Text Rendering. The evaluator receives the complete story specification together with the generated trajectory and assigns an absolute score from $1$ to $5$ for each applicable dimension. Scores are assigned at the trajectory level rather than by enforcing a potentially incorrect one-to-one alignment between generated images and specified shots.

The evaluation rubric explicitly distinguishes consistency from degenerate repetition. Reusing nearly identical images may preserve identity but does not demonstrate visual memory or scene differentiation. Similarly, a sequence that reuses essentially the same environment throughout the story is capped at $2$ for Scene Memory. Visual Fidelity and Aesthetic Quality are judged relative to the visual style requested by the specification, so intentionally stylized or non-photorealistic outputs are not penalized for deviating from photographic appearance.

\subsection{Evaluator Prompts}
\label{app:evaluator_prompts}

For reproducibility, we provide the complete prompts used by the GPT-6 trajectory-level evaluators. Generated frames are supplied together with the textual prompt, and the evaluator is required to return structured JSON. We use three specialized evaluators: the primary trajectory evaluator for Character Identity, Scene Memory, and Visual Fidelity; a second evaluator for Visual Memory and Aesthetic Quality; and a third evaluator dedicated to Text Rendering. 

\begin{promptbox}{Primary trajectory evaluator: Identity, Scene, and Visual Fidelity}
You are an expert evaluator of AI-generated visual stories (storyboards, picture books, comics).

You will be given (1) a STORY SPEC that every system was asked to render, and (2) the frames ONE system produced, in the order it produced them.

CRITICAL FRAMING RULES
- The number of frames may differ from the number of shots in the spec. These systems plan their own storyboards. Do NOT reward or penalise a system merely for frame count, and do NOT assume frame k corresponds to shot k. Work out the correspondence yourself from the content.
- Judge the frames as a SEQUENCE, not as isolated pictures.
- Judge craft against the story's DECLARED art style. If the spec asks for flat colour, watercolour, greyscale or black-and-white manga, then correctly following that is GOOD. Never reward photorealism or saturated colour for its own sake.
- Ignore any watermark, site header or page furniture.
- Do not give courtesy scores. Use the full 1-5 range. 3 means "acceptable but with clear defects".

Score each axis as an INTEGER 1-5.

1. character_identity
Is each named cast member the SAME individual in every frame where they appear (face/muzzle structure, body proportions, hair or fur colour, clothing and its details, accessories)? Pose, expression, camera distance and lighting are ALLOWED to change and must not be penalised.
5 = a reader would never doubt it is the same character.
4 = one or two small slips, e.g. an accessory changes.
3 = recognisable but clearly drifting.
2 = frequently reads as a different character.
1 = essentially a new character each frame, or the species/kind changes.

2. scene_memory
When the story RETURNS to a location after visiting other locations, does that location read as the SAME place (architecture, layout, palette, signage, landmark props)? AND do DIFFERENT locations stay clearly DISTINCT from each other?
5 = revisits are unmistakably the same place AND distinct locations are clearly different.
3 = revisits are only roughly similar, or some distinct locations blur together.
1 = revisits look unrelated, OR everything is one generic backdrop.
HARD RULE: if the system renders essentially one background for the whole story, score at most 2. Do not score merely-similar generic vegetation or a blank studio backdrop as a successful revisit.
Do not score above 3 just because two frames share a location NAME; the depicted place must actually match.

3. visual_quality
Craft and coherence within the declared style: anatomy, composition, rendering, freedom from artifacts and from garbled/nonsense lettering.
5 = clean and controlled.
3 = usable with visible defects.
1 = broken.

PROCEDURE
First identify the locations depicted across the sequence and compare frames corresponding to scene revisits before assigning scene_memory.
Then score each reported axis with a brief justification citing specific frames.

Return JSON only.
\end{promptbox}

The structured output additionally contains a one-sentence justification for each axis, together with \texttt{revisit\_\allowbreak analysis}, \texttt{degenerate}, and \texttt{note}. The main comparison uses Character Identity, Scene Memory, and Visual Fidelity from this evaluator.

\begin{promptbox}{Visual Memory and Aesthetic Quality evaluator}
You are judging an AI visual story on TWO axes only.

You get: the declared STYLE, a short CAST/LOCATION spec, and the frames
ONE system produced (in order). Frame count may differ from the spec.
Do NOT assume frame k is shot k.

1. visual_memory
Did LATER frames reuse visual decisions that were first COMMITTED IN EARLIER
GENERATED FRAMES, not merely restated in the text?
Examples of memory: the exact mole placement, a specific shop-sign layout,
a unique knit pattern, the same window frame geometry after the camera left
and came back.
HARD RULES
- Matching a user reference photo is NOT memory. That is image conditioning.
  If the face is locked to a reference but locations/outfits/backgrounds
  are re-rolled or destroyed, score at most 2.
- Copying one panel into every frame is NOT memory. Set copy_collapse=1
  and score visual_memory = 1.
- Independent T2I that only follows the written description, with details
  re-rolling each frame, should score 2-3 even if each still looks pretty.
5 = later frames clearly remember earlier generated choices after a change
    of location or pose.
3 = mostly follows the written spec; little extra carry-over.
1 = new character/place each time, or one image repeated.

2. style_aesthetics
Beauty and craft INSIDE the declared style. If the spec asks for manga,
watercolour, greyscale, film grain, or hard flash, following that is GOOD.
Do NOT reward photorealism for its own sake.
MUST penalise: white speckle/glitter overlays, decorative polaroid/phone
chrome, garbled lettering, broken anatomy, one wallpaper reused as every
background.
5 = clean, attractive, on-style.
3 = usable, visible defects.
1 = broken or buried in artifacts.

Fill memory_analysis and style_analysis FIRST. Return JSON only.
\end{promptbox}

\begin{promptbox}{Text Rendering evaluator}
You are judging TEXT RENDERING.

STEP 1 -- list text frames
A frame counts if it has painted story text: shop signs, neon, price tags,
posters, magazines, menus, titles, speech balloons, handwritten notes.
Does NOT count: watermarks, site headers, Polaroid/Instagram chrome, EXIF
junk, pure texture that is not writing.
If ZERO frames qualify: set n_text_frames=0, text_frames="", text_seen=NONE,
and set text_rendering=0 and text_match=0. Do not invent a 1-5 score.

STEP 2 -- score only those frames (when n_text_frames >= 1)

1. text_rendering -- glyph quality on text frames only
Are the characters real and readable (correct Chinese / English strokes),
not melted fake radicals?
5 = almost all text frames are clean and readable.
3 = mixed: some words OK, others broken.
1 = writing is present but almost all garbled.

2. text_match -- wording on text frames only
If the spec names specific words, are they present and correct?
If the spec only wants generic signage, plausible on-theme text in the
right language is enough; random English on a Chinese street scores low.
5 = requested or clearly on-theme words are correct.
3 = right language / vibe, wording drifts.
1 = wrong language or nonsense.

Return JSON only.
\end{promptbox}

\section{Baseline Implementation Details}
\label{app:baselines}

\subsection{Evaluation Paradigms}
\label{app:baseline_paradigms}

We evaluate two families of long-form generation systems. \textit{Agent-Planned} methods decompose the story-level specification into a sequence of image-generation instructions and synthesize each image independently. This group includes StoryDiffusion, Qwen-Image, UniSpace, HiDream, Lance-3B, and Show-o2-7B. \textit{Interleaved} methods instead receive the story-level instruction and generate text and images autoregressively within a shared multimodal trajectory. This group includes BAGEL-7B, SenseNova-U1.5, ILLUME-X, Emu3.5, SEED-Story, and UniLongGen$^{*}$.

The two groups differ in how historical information is propagated. Agent-Planned systems rely on the information explicitly included in each image-generation instruction, whereas native interleaved systems can condition subsequent generation on the multimodal states already produced within the trajectory. All systems are evaluated using the same LongMIG-Bench story specifications and the evaluation protocols described in Appendix~\ref{app:evaluation}.

\subsection{Agent-Planned Baselines}
\label{app:agent_baselines}

For Agent-Planned baselines, each story is represented as a sequence of per-image generation instructions derived from the same underlying structured specification. We use GPT-6 as a shared planner for all Agent-Planned methods to decompose each story into an ordered shot plan and generate the corresponding per-image captions. This ensures that differences across methods primarily reflect their image-generation capabilities rather than differences in planning or prompt construction. Images are then generated independently at each step rather than being inserted into a shared autoregressive multimodal context. For the image-conditioned track, the provided user reference image is supplied whenever the corresponding method supports reference-conditioned generation.

We include strong general-purpose image generators as well as methods designed for multi-image or story generation. StoryDiffusion provides a representative story-generation baseline, while Qwen-Image, UniSpace, HiDream, Lance-3B, and Show-o2-7B provide strong per-image generation capability under the agent-planned formulation.

\subsection{Native Interleaved Baselines}
\label{app:interleaved_baselines}

Native interleaved baselines are evaluated in their text-image generation mode, where the model receives the complete story-level instruction and produces a multimodal autoregressive trajectory. We evaluate BAGEL-7B, SenseNova-U1.5, ILLUME-X, Emu3.5, and SEED-Story. Unlike the Agent-Planned setting, generated images remain part of the evolving multimodal trajectory according to each model's native inference interface.

For all methods, generated character-reference or auxiliary reasoning images that are not part of the final story are excluded from story-panel metrics. The evaluator receives only the resulting story panels together with the same ground-truth LongMIG-Bench specification.

\subsection{UniLongGen Same-Backbone Reimplementation}
\label{app:unilonggen}

To separate the effect of visual-memory management from backbone capability, we additionally reimplement the training-free context-curation strategy of UniLongGen on the same unified multimodal backbone used by our method, denoted as UniLongGen$^{*}$. This comparison removes the large variation in image-generation quality and multimodal capability that otherwise exists across different unified models.

UniLongGen$^{*}$ follows the historical visual-context selection strategy of UniLongGen while retaining our underlying model weights. In contrast, our method constructs persistent visual references, predicts the current scene state, and retrieves a bounded set of scene-relevant historical generative states. The comparison therefore focuses on the effect of the visual-memory strategy under a shared backbone.

\section{Additional Efficiency and Scalability Analysis}
\label{app:efficiency}

We provide additional analysis of the computational behavior of our visual-memory mechanism as the generation horizon increases. Unless otherwise specified, Ours and Full History use identical model weights, decoding settings, and hardware; the only difference is whether historical generative visual states are selectively retrieved or all retained for image synthesis. Both settings preserve the complete semantic visual history for multimodal understanding.

\subsection{Efficiency across Generation Horizons}
\label{app:efficiency_horizon}

We first evaluate eight long-form stories, each extended to 32 shots, at increasing generation horizons. Inference uses the same two-node setup for both policies, with one AR node (TP$=8$) and one image-generation node (TP$=8$), and 30 denoising steps for each generated image.

\begin{table}[t]
\centering
\caption{\textbf{Efficiency across increasing generation horizons.}
Ours and Full History use identical model weights and generation settings. Sec./Image is the mean wall-clock time per image over the first $h$ images; Last Prefill is the AR prefill time of the $h$-th image.}
\label{tab:efficiency_scaling}
\small
\setlength{\tabcolsep}{5.0pt}
\begin{tabular}{ccccc}
\toprule
& \multicolumn{2}{c}{Sec./Image $\downarrow$}
& \multicolumn{2}{c}{Last Prefill (s) $\downarrow$} \\
\cmidrule(lr){2-3}\cmidrule(lr){4-5}
Horizon & Full History & Ours & Full History & Ours \\
\midrule
8  & 27.8  & 26.4 & 6.8  & 4.9 \\
16 & 49.3  & 37.6 & 20.7 & 9.8 \\
24 & 73.8  & 46.5 & 41.4 & 14.0 \\
32 & 102.9 & 54.9 & 69.0 & 19.4 \\
\bottomrule
\end{tabular}
\end{table}

As shown in Table~\ref{tab:efficiency_scaling}, the computational gap increases steadily with generation horizon. At 32 images, selective visual-memory retrieval reduces average generation time from $102.9$ to $54.9$ seconds per image, while the AR prefill time of the final image decreases from $69.0$ to $19.4$ seconds. The two policies are relatively close at short horizons, where fixed image-generation cost dominates, but diverge as historical visual states accumulate. Ours is faster on every story available under both policies at every horizon of 16 images or more.

\subsection{Context and Latency Growth}
\label{app:context_scaling}

\begin{figure}[t]
\centering
\includegraphics[width=1\linewidth]{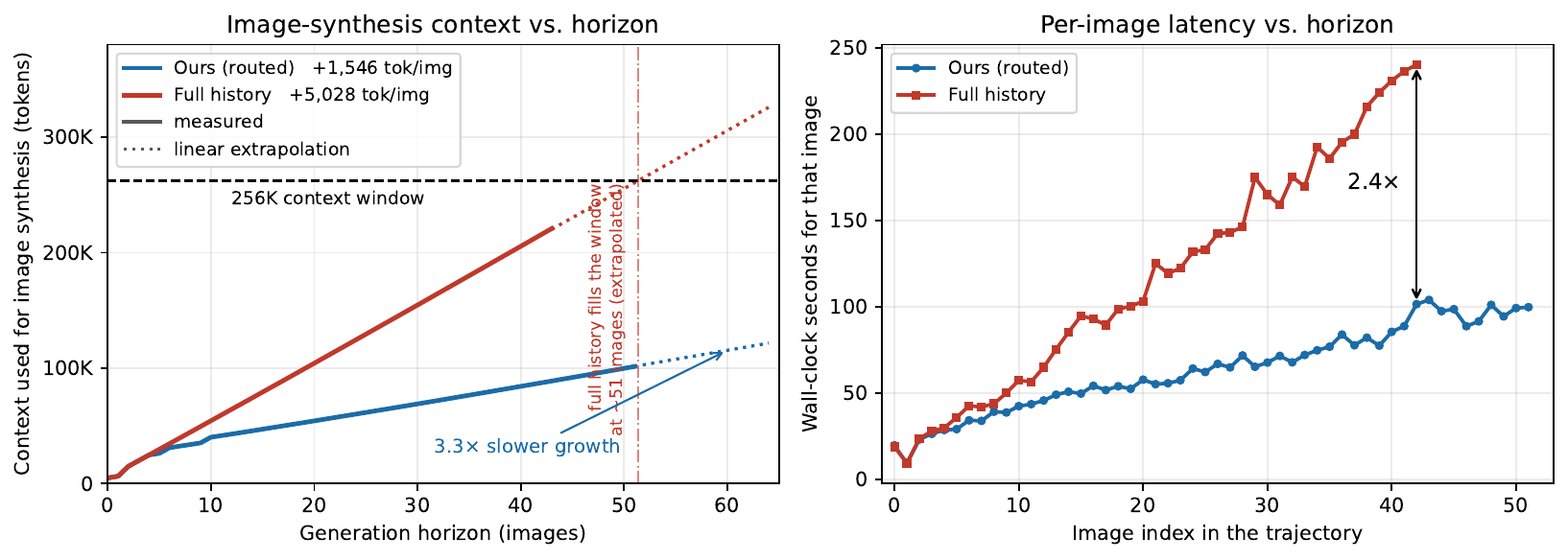}
\caption{\textbf{Scaling of image-synthesis context and generation latency.}
Left: selective visual-memory retrieval substantially slows context growth compared with Full History. Right: the corresponding per-image latency gap increases with generation horizon. Solid curves denote measured trajectories, while dotted extensions indicate linear extrapolation.}
\label{fig:horizon_scaling}
\end{figure}

To isolate the effect of historical visual conditioning, we additionally trace a fixed long-form generation under both policies. Figure~\ref{fig:horizon_scaling} reports the image-synthesis context and per-image wall-clock time as the trajectory grows.

In this controlled trace, the image-synthesis context grows by approximately $1{,}546$ tokens per additional image with our routing policy, compared with $5{,}028$ tokens per image under Full History. The corresponding wall-clock growth is approximately $1.5$ versus $5.5$ seconds per additional image. Thus, the marginal latency introduced by extending the trajectory grows substantially more slowly with selective visual-memory retrieval. Because the denoising schedule is fixed, the attention cost of image synthesis scales linearly with this context; at 32 images the context is $71.9$K tokens under our policy versus $164.6$K under Full History, a $2.3\times$ reduction.

\section{Failure Cases and Limitations}
\label{app:limitations}

Despite the improvements in long-horizon identity and scene consistency, several failure modes remain. We summarize representative cases and clarify the scope of the conclusions supported by our experiments.

\subsection{Fine-Grained Object Continuity}
\label{app:limit_objects}

Our visual-memory mechanism operates primarily at the level of complete historical images and persistent visual references. It does not explicitly bind fine-grained object instances across the trajectory. Consequently, small carried objects, accessories, garment details, or objects placed within a scene can still disappear, change appearance, or be replaced after multiple generations.

\subsection{Dependence on Explicit Scene-State Prediction}
\label{app:limit_scene_state}

The current retrieval mechanism depends on an explicit discrete scene state. During training, scene annotations are obtained from the structured training data, whereas at inference time the model generates the scene state within its textual reasoning trajectory and the memory controller parses this prediction to determine the relevant historical visual states.

This creates a coupling between the model's generated reasoning format and the memory mechanism. If the model fails to emit a valid scene state, changes the output format, or predicts the wrong scene, retrieval may expose irrelevant historical visual states or fail to recover the intended establishing view. A representation-level scene state predicted directly from hidden features would remove this dependence on textual serialization and is a promising direction for future work.

\subsection{Remaining Long-Horizon Generation Failures}
\label{app:limit_generation}

Long-horizon consistency is improved but not perfect. We observe several recurring failure modes: identity drift in crowded multi-character scenes, imperfect recovery of fine scene details after very long gaps, repeated compositions in design-oriented sequences, and occasional degradation in rendered text. Self-rollout reduces exposure mismatch but does not eliminate error accumulation when an early generation contains a substantial visual error.

These failures become more visible as the trajectory grows because generated outputs become part of the context for subsequent generation. Our method mitigates this accumulation by controlling which visual states are reused, but it cannot correct every error already encoded in the generated history.

Overall, our results demonstrate that visual-memory management substantially extends the usable horizon of native interleaved generation, but it does not eliminate all forms of autoregressive error accumulation. More structured object-level memory, representation-level scene tracking, and stronger adaptation to self-generated histories remain promising directions for further improving long-form multimodal generation.

\end{document}